\documentclass[accepted]{uai2026} 
\usepackage[american]{babel}
\usepackage{subcaption}
\usepackage{natbib} 
\usepackage{mathtools} 
\usepackage{booktabs} 
\usepackage{tikz} 
\usepackage[linesnumbered,lined, ruled, commentsnumbered]{algorithm2e}
\usepackage{makecell}
\usepackage{amssymb}
\newcommand{\algname}{Maximally Robust Satisficing Bayesian Optimization}
\newcommand{\algabbrev}{MRSBO}

\title{Maximally Robust Satisficing Bayesian Optimization}

\author[1]{\href{mailto:<samuli.kinnunen@helsinki.fi>?Subject=Maximally Robust Satisficing Bayesian Optimization}{Samuli~Kinnunen}{}}
\author[1]{Petrus~Mikkola}
\author[2]{Antti~Niskanen}
\author[1]{Arto~Klami}
\affil[1]{%
    Department of Computer Science\\
    University of Helsinki\\
    Finland
}
\affil[2]{%
    ASM International N.V.
  }
  
\begin{document}

\maketitle
\begin{abstract}
Many design tasks can be cast as black-box function optimization, enabling use of Bayesian optimization to find an ideal design with minimal number of trials. However, often we do not actually need the optimum but instead a sufficiently good solution is enough, for instance a material that is durable enough for its intended use.
  In most cases there are multiple satisfactory solutions, forming a superlevel set of the function, raising a key question of which one to prefer. We answer this by explaining why robustness to input perturbations that may occur when the solution is deployed is a good criterion and by introduce a  Bayesian optimization method that efficiently finds satisficing solutions that are robust to maximally large perturbations. In contrast to previous works, we assume the inputs can be accurately controlled during optimization, but will be perturbed after the deployment.
\end{abstract}

\section{Introduction}\label{sec:intro}
Bayesian optimization (BO) is a sample-efficient framework for optimizing expensive black-box functions, applied in diverse domains such as hyperparameter tuning \cite{cho2020basic}, materials discovery \cite{chitturi2024targeted}, chemical synthesis \cite{shields2021bayesian}, and robotic control \cite{lechuz2024bayesian}. 
For the standard problem of finding the global optimum, $x^* = \arg \max_{x} f(x)$, there are highly polished tools. However, for many design tasks where the evaluations are extremely costly, e.g. due to involving production and testing of a new material, even a sample-efficient search may still be difficult in practice.

Often we do not, however, need the global optimum $x^*$, but already a sufficiently good solution that meets a given minimum quality, $f(x^*) \ge t$, is enough. 
For instance, a subcontractor may be asked to produce a component that achieves a certain level of durability, reducing the difficulty of the problem.
We call such a solution \emph{satisficing}, following the nomenclature introduced by Herbert Simon \citep{simon1955behavioral} and subsequently used e.g. in operations research and decision-making \citep{reverdy2016satisficing, long2023robust, saday2025robust}.

The set of all satisficing solutions forms a superlevel set $S_t = \{x : f(x) \ge t\}$ of the function. In principle any solution within $S_t$ would work, but outside of idealized conditions they are not alike. One would intuitively steer away from solutions close to the threshold of $f(x)=t$ and prefer a solution with a higher value, but we need a more formal criterion. We argue that robustness to input noise or variation is a good one: A solution is good if $f(x^*)$ remains satisficing when $x^*$ is perturbed.

\begin{figure}[t]
    \centering
    \includegraphics[width=1\linewidth]{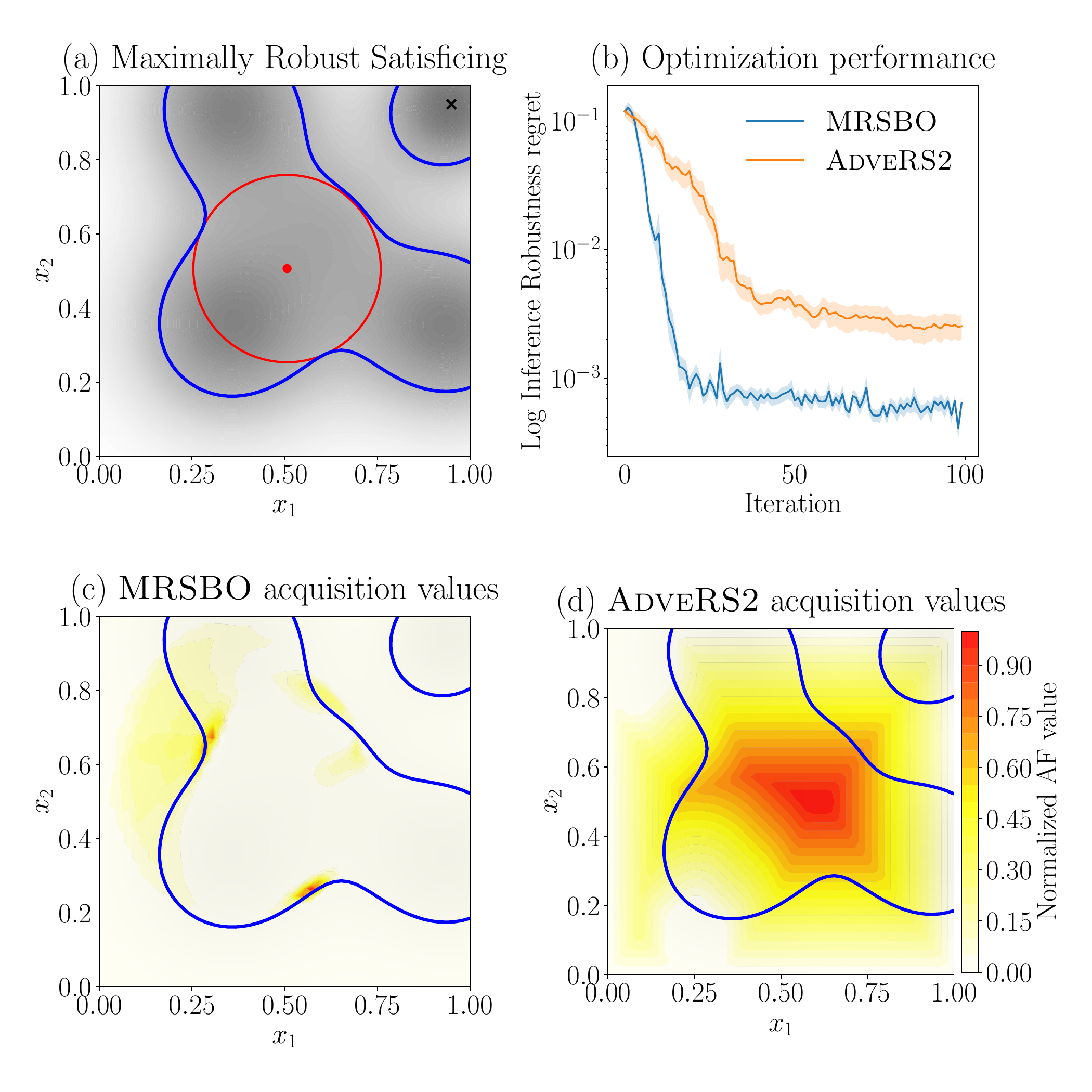}
    \caption{{\bf(a)} The maximally robust satisficing solution (red) is a point that can be modified the most while still staying within the superlevel set defined by the threshold that needs to be satisfied (blue). Note that the global maximum (black cross) can be arbitrarily far.
    {\bf(b)} Our MRSBO method drastically outperforms the
    best previous method (AdveRS2) adapted for our setup; it also optimizes for the robustness radius, but assumes input perturbations already during optimization.
    {\bf(c-d)} Acquisition functions for the two methods during late stages of optimization. MRSBO explores the parts of the level set that are informative of the robustness radius, whereas AdveRS2 wastes evaluations for learning the target within the superlevel set.}
    \label{fig:placeholder}
\end{figure}

We draw motivation from applications where the inputs can be accurately controlled in the design (optimization) phase but are unreliable after deployment. As a concrete example, consider a cake recipe designed by professionals that measure the ingredients accurately and use consistent kitchen equipment. When commoners bake the cake following the recipe, they will not measure their tablespoons exactly and their ovens do not maintain the temperature perfectly. What ultimately matters is how good \emph{their} cake tastes, and hence a recipe robust to this kind of deviations is a better one. The same principle is seen in many scientific applications: manual procedures and environmental factors like humidity introduce variability in laboratory chemistry \citep{aldeghi2021golem}, sim-to-real transfer \citep{doersch2019sim2real} is a whole subfield studying specifically how to train reliable real-world models using clean simulation environments, and in design and manufacturing \citep{daulton2022robust} fabrication tolerances constraint the realizability of designs.
Often the solution that is optimal under noiseless inputs can perform poorly or even lead to unsafe behavior under perturbations \cite{oliveira2019bayesian}.

To address this need, we introduce a new problem: Given a threshold $t$ that determines the minimum acceptable value for the function, find a solution $x^*$ that is robust to \emph{maximally large} input deviations. During optimization we assume $f(x)$ can be evaluated for arbitrary $x$, but the solution is eventually deployed in an environment where the inputs are perturbed by an unknown noise. That is, instead of $f(x^*)$ we observe $f(x^* + \delta)$.
The task is illustrated in Figure~\ref{fig:placeholder}, showing also how our method 
maximizing the information gain about the maximal robustness radius solves the problem efficiently by primarily evaluating the function around selected parts of the level set, not wasting evaluations to learn the function accurately within the superlevel set.

While our problem setup is new, it is closely related to the concept of \emph{Robust Bayesian Optimization (RBO)}, generally referring to optimization methods finding solutions that remain effective under input perturbations.
Previous RBO methods can be broadly categorized into two types. \emph{Expectation Robustness} \citep{yang2023efficient,frohlich2020noisy}
optimizes expected performance under a known perturbation distribution $p(\delta)$, 
$
x^* = \arg \max_{x} \mathbb{E}_{\delta \sim p(\delta)} \big[ f(x + \delta) \big],
$
with \emph{Distributionally Robust Optimization} \citep{kirschner2020distributionally,shapiro2023bayesian} accounting also for uncertainty in the perturbation distribution. 
These methods ensure the expected performance to be good, but there are \emph{no} guarantees for the solution to satisfy $f(x)>t$ for any $x$.

\emph{Worst-case Robustness} \citep{bertsimas2010robust,bogunovic2018adversarially} is closer to our setup, searching for a solution that performs optimally under the most adverse  perturbation as
$
x^* = \arg \max_{x} \min_{\delta} f(x + \delta).
$
Most methods in this family, like StableOpt by \cite{bogunovic2018adversarially}, directly optimize for the worst-scare performance, but recently \cite{saday2025robust} connected the worst-case robustness to satisficing solutions. They also maximize the magnitude of the perturbations that can be allowed while ensuring the solution remains above the threshold $t$. However, there is a fundamental difference in the learning setup: Similar to other worst-case robustness methods, \cite{saday2025robust} assume already the optimization is done in an adversarial environment. This is in stark contrast to our setup where the inputs are clean during optimization. Their method can be used to solve our problem by artificially inducing perturbations during optimization, but this requires assuming a maximum level of perturbation. Figure~\ref{fig:placeholder} shows how this approach in practice fails even when given oracle information about the maximum perturbation, with substantially worse convergence, poor final regret, and wasteful evaluations.

\textbf{Contributions.} Our main contributions are: (1) We formalize Maximally Robust Satisficing Bayesian Optimization problem setup for learning solutions that remain sufficiently good under maximal input perturbation.
(2) We provide practical algorithm \algabbrev{ for }solving it, based on the information gain about the maximally robust solution with an efficient construction for the acquisition function.
(3) We demonstrate that \algabbrev{} vastly outperforms robust and non-robust alternatives on a variety of synthetic and real-world robust optimization problems.

\section{Background}
\subsection{Bayesian Optimization Framework}
Bayesian Optimization, in its standard form, is a global optimization method for solving 
\begin{equation}
    x^* = \arg\max_{x\in \mathcal{X}} f(x),
\end{equation}
where $\mathcal{X}\subseteq \mathbb{R}^d$ is a $d$-dimensional design space and $f: \mathcal{X} \rightarrow \mathbb{R}$ is an unknown function that is expensive to evaluate.
To optimize the function, we can only evaluate inputs $x$ to produce an evaluation $y = f(x)$. BO algorithm learns cheap surrogate model, typically a Gaussian process, that approximates the real function and selects the next evaluation by maximizing an acquisition function
to identify the evaluation that maximally benefits the optimization process. 
\textbf{Surrogate model} approximates the true objective function $f$. A Gaussian process (GP) \cite{williams1995gaussian} is the most commonly used surrogate model, providing a distribution over function values $f(x)$ characterized 
by a mean function $\mu(x)$ and a covariance (kernel) function $k(x,x')$. 
Given observations $D_t = \{(x_i, y_i)\}_{i=1}^t$, the posterior predictive distribution at a test point $x$ is Gaussian with mean $\mu_t(x) = k_t(x)^\top (K + \sigma_{\epsilon}^2 I)^{-1} y$ and variance $\sigma_t^2(x) = k(x,x) - k_t(x)^\top (K + \sigma_{\epsilon}^2 I)^{-1} k_t(x),$, where $k_t(x) = [k(x_i, x)]_{i=1}^t$, $K = [k(x_i, x_j)]_{i,j=1}^t$ is the kernel matrix of the training inputs, $y = [y_1, \dots, y_t]^\top$, and $\sigma_{\epsilon}^2$ denotes the observation noise variance.

\textbf{Acquisition function (AF)} $\alpha(x)$ scores the value of evaluating an input point $x$. At each Bayesian optimization iteration, the next query point is selected by solving $x_{t+1} = \arg\max_{x \in \mathcal{X}} \alpha(x)$. Popular single-objective acquisition functions include Upper Confidence Bound (UCB), Expected Improvement (EI) \cite{jones1998efficient} and Max-value Entropy Search (MES) \cite{wang2017max}.

\section{Problem Definition}

We consider design tasks where a solution $x^*$ is determined in controlled settings, but after fixing the design it is deployed in an environment where every input $x \in \mathcal{X}$ may experience an unknown perturbation $\delta$ whose magnitude is bounded by an unknown robustness radius $r>0$.

For any $r$, the admissible perturbations are collected in
\[
B_r=\left\{\delta \in \mathbb{R}^d : \|\, \delta \,\| < r \,\right\},
\]
and we define as satisficing the $x \in \mathcal{X}$ that remain satisfactory under all admissible perturbations: the worst--case function value stays above a given threshold $t \in \mathbb{R}$ as
\[
\inf_{\delta \in B_r} f(x+\delta) \;\ge\; t.
\]
To find the maximally robust solution, we define
the \emph{objective function} $\rho(x)$ as the maximum robustness radius $r$ such that all perturbed inputs yield satisficing values:
\begin{align}
    \rho(x) = \sup \left\{ r > 0 : \inf_{\delta \in B_r} f(x + \delta) \ge t \ \text{and} \ x + \delta \in \mathcal{X}\right\}. \label{eq:rho_objective_definition}
\end{align}
This immediately gives the final learning task that provides both the optimal solution $x^*$ and the corresponding robustness radius $r^*$ as
\begin{align}
    x^* = \arg\max_{x \in \mathcal{X}} \rho(x),\label{eq:x_star_is_rho}\\ 
    r^* = \rho(x^*).
\end{align}

Our \emph{objective} is identical to the one recently proposed by \cite{saday2025robust}, but the \emph{learning setup} is different. We assume exact evaluations $f(x)$ during optimization, matching the motivational example cases, whereas they assume the inputs are perturbed also during optimization (and that the perturbed inputs can, perhaps unrealistically, be observed for learning the proxy).

\section{Method}
This section introduces the \algabbrev{} method for solving the above problem, summarized in Algorithm~\ref{alg_BO}. 
It follows a standard Bayesian optimization protocol: 
At each iteration, candidate maximizers are sampled (Line 3), and the acquisition function is optimized to select the next evaluation point (Line 4). The proxy is  updated with the new observation, and we continue until exhausting an evaluation budget $T$.

We here design an information-theoretic acquisition function and explain how it can be computed efficiently. The acquisition maximization step can be carried out using standard gradient-based optimization methods, e.g., as discussed by \cite{wilson2018maximizing}; we leave description of this standard procedure in Appendix~\ref{app_implementation}.

\begin{algorithm}[t]
\KwIn{Design space $\mathcal{X}$, GP prior $(\mu(x), k(x, x'))$, acquisition function $\alpha(x)$, budget $T$}
\KwOut{Recommended point $\tilde{x}^*$}
Initialize data $\mathcal{D}_0$\\
\For{$t = 1,\dots,T$}{
    $\{(x^{*,(s)}, r^{*,(s)})\}_{s=1}^n \gets \textsc{SampleContinuous}$ \\
    $x_t \gets \underset{x' \in \mathcal{X}}{\arg\max}\ 
    \alpha\!\left(x', \{(x^{*,(s)}, r^{*,(s)})\}_{s=1}^n, \mathcal{D}_{t-1}\right)$\\
    $y_t \gets f(x_t) + \epsilon_t$\\
    $\mathcal{D}_t \gets \mathcal{D}_{t-1} \cup \{(x_t, y_t)\}$\\
}
$\tilde{x}^* \gets \arg\max_{x \in \mathcal{X}} \tilde{\rho}(x)$
\caption{\algname{} (\algabbrev)}
\label{alg_BO}
\end{algorithm}

\subsection{Information gain about $(x^*,r^*)$}\label{sec_information_gain}

We propose an acquisition strategy that maximizes the information gain about the maximum robustness radius $r^*$ and its corresponding robust maximizer $x^*$. Following the symmetric mutual information formulation \citep{hernandez2014predictive, wang2017max}, this approach avoids direct entropy computation of the maximizer and robustness radius, which are difficult quantities even to sample.

Formally, the acquisition function for candidate $x$ is 
\begin{align}
\alpha(x)
    &= I(\{y,x\}; \{x^*, r^*\}) \nonumber \\ 
    &= I(\{x^*, r^*\};\{y,x\}) \nonumber \\ 
    &= H(p(y \mid x,D_t)) \label{eq_IG_entropy}\\
    &- \mathbb{E}_{p(x^*, r^*|D_t)} \bigl[H(p(y \mid x,D_t, x^*, r^*))\bigr] \nonumber
\end{align}

The first term of Eq. \ref{eq_IG_entropy}, $H(p(y \mid x, D_t))$, is the entropy of a Gaussian random variable and admits the analytic form
\begin{align}
H\bigl(p(y \mid x, D_t)\bigr)
    = \frac{1}{2} \log \bigl(2\pi e \sigma_t^2\bigr),
\end{align}
where $\sigma_t^2$ denotes the posterior predictive variance at $x$.

The expectation in the second term of Eq.~\ref{eq_IG_entropy} is taken with respect to the joint posterior of the robust maximizer and robustness radius, and is estimated using Monte Carlo sampling. We next discuss the approximation of the conditional entropy, $H(p(y \mid x,D_t, x^*, r^*))$, starting with a noiseless case for communicating the core idea even though in practice we always assume noisy observations.

\paragraph{Noiseless conditional entropy.}
In the absence of observation noise ($\sigma_{\epsilon}^2=0$), the computation can be separated into two cases based on the distance $d(x,x^*)$. If the candidate $x$ is within the robustness radius
($
d(x,x^*) \le r^*
$)
then, by definition of robustness, we are guaranteed to have $f(x) \ge t$. When
$
d(x,x^*) > r^*,
$
we do not have any information about the values of $f(x)$ unless making additional assumptions about the function.
$f(x)$ can still be arbitrarily high since the superlevel set can take any form. Hence, the conditional entropy can be expressed as:
\begin{align}
    &H(p(y \mid x,D_i, x^*, r^*)) \nonumber \\
    &\approx\begin{cases}
        \log(\sqrt{2\pi e}\,\sigma Z) + \frac{\alpha \varphi(\alpha)}{2Z}, & \text{if } d(x, x^*) \leq r^*, \\
        \frac{1}{2} \log (2\pi e \sigma_t^2), & \text{otherwise},
    \end{cases}
    \label{eq_conditional_entropy}
\end{align}
where $\alpha = (a - \mu)/\sigma_t$ is the standardized lower truncation point, with $a$ being the lower bound. Z and $\varphi$ are defined as
\[
Z = 1-\Phi(\alpha)
\qquad
\varphi(\xi) = \frac{1}{\sqrt{2\pi}}\exp\left(-\frac{1}{2}\xi^2\right),
\]
where $Z(x) = 1 - \Phi\!\bigl(\alpha\bigr)$ can be computed using the standard normal cumulative distribution function $\Phi$.

For smooth functions we could attempt lower-bounding the function also outside of the robustness radius: For $x$ just outside the robustness radius, we necessarily have $f(x)$ close to $t$.
Exploiting this in practice, however, would require additional assumptions about the smoothness of the function, which are technically challenging to specify correctly \citep{lederer2019uniform} and yield little empirical benefit, as shown in Appendix~\ref{appendix_conditional_with_lipschitz}. We therefore recommend using the above conditional
entropy approximation.

\paragraph{Noisy conditional entropy.}
With observation noise, the guarantee that $f(x) > t$ for $d(x, x^*) \leq r^*$ no longer holds, but the distance still provides information. For computing the entropies, we adapt the approach of \cite{takeno2020multi} proposed for noisy truncation in multi-fidelity BO context. The conditional predictive density can be expressed as
\begin{align}
p(y \mid x, D_t, f > t) = Z(y)\, p(y \mid x, D_t),
\end{align}
where $p(y \mid x, D_t)$ is the posterior predictive density and
\[
Z(y) = \frac{p(f>t \mid y, x, D_t)}{p(f>t \mid x, D_t)}.
\]
The denominator corresponds to a Gaussian tail probability, while the numerator is computed from the conditional distribution
$
f \mid y, x, D_t \sim \mathcal{N}(u, s^2),
$
with
\[
u = \mu + \frac{\sigma_f^2}{\sigma_f^2 + \sigma_{\epsilon}^2}(y - \mu),
\qquad
s^2 = \sigma_f^2 - \frac{\sigma_f^4}{\sigma_f^2 + \sigma_{\epsilon}^2}.
\]
Finally, the conditional entropy $H(p(y \mid x,D_t, x^*, r^*))$
is approximated via Monte Carlo samples $\{y_i\}_{i=1}^N$ as
\begin{align} 
-\frac{1}{N}\sum_{i=1}^N \log p(y_i \mid x, D_t, f>t) & \text{  if } d(x, x^*) \leq r^*, \nonumber \\
\frac{1}{2} \log (2\pi e \sigma_t^2), & \quad \text{ otherwise}.\nonumber 
\end{align}

\subsection{Sampling from $p(x^*, r^*|D_t)$}\label{subsec:sampling}

Evaluation of the acquisition function requires computing expectations with respect to the posterior  $p(x^*, r^* \mid \mathcal{D}_t)$. Since this distribution is generally intractable, we approximate the
expectation using Monte Carlo integration. This reduces the problem to efficiently sampling pairs
$(x^*, r^*)$.

Following \cite{hernandez2014predictive, hernandez2016predictive}, we approximate samples from the GP posterior using Random Fourier Features (RFF). This yields differentiable sample paths for the sampling procedure in Algorithm~\ref{alg_continuous_sampling}. 

\begin{figure}[t]
    \centering
    
    \begin{subfigure}{0.48\linewidth}
        \centering
        \includegraphics[width=\linewidth]{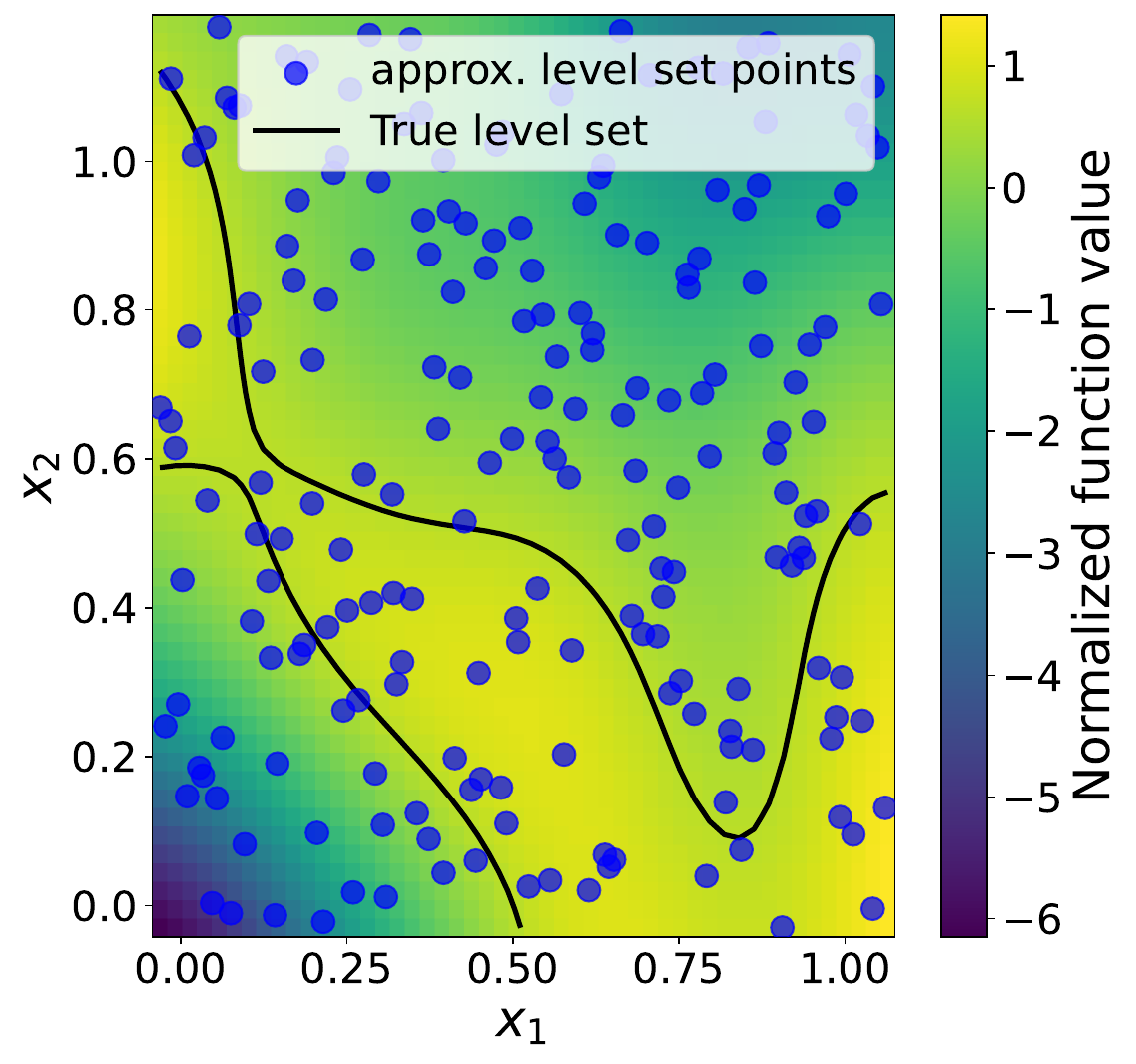}
        \caption{}
        \label{fig:init}
    \end{subfigure}
    \hfill
    \begin{subfigure}{0.48\linewidth}
        \centering
        \includegraphics[width=\linewidth]{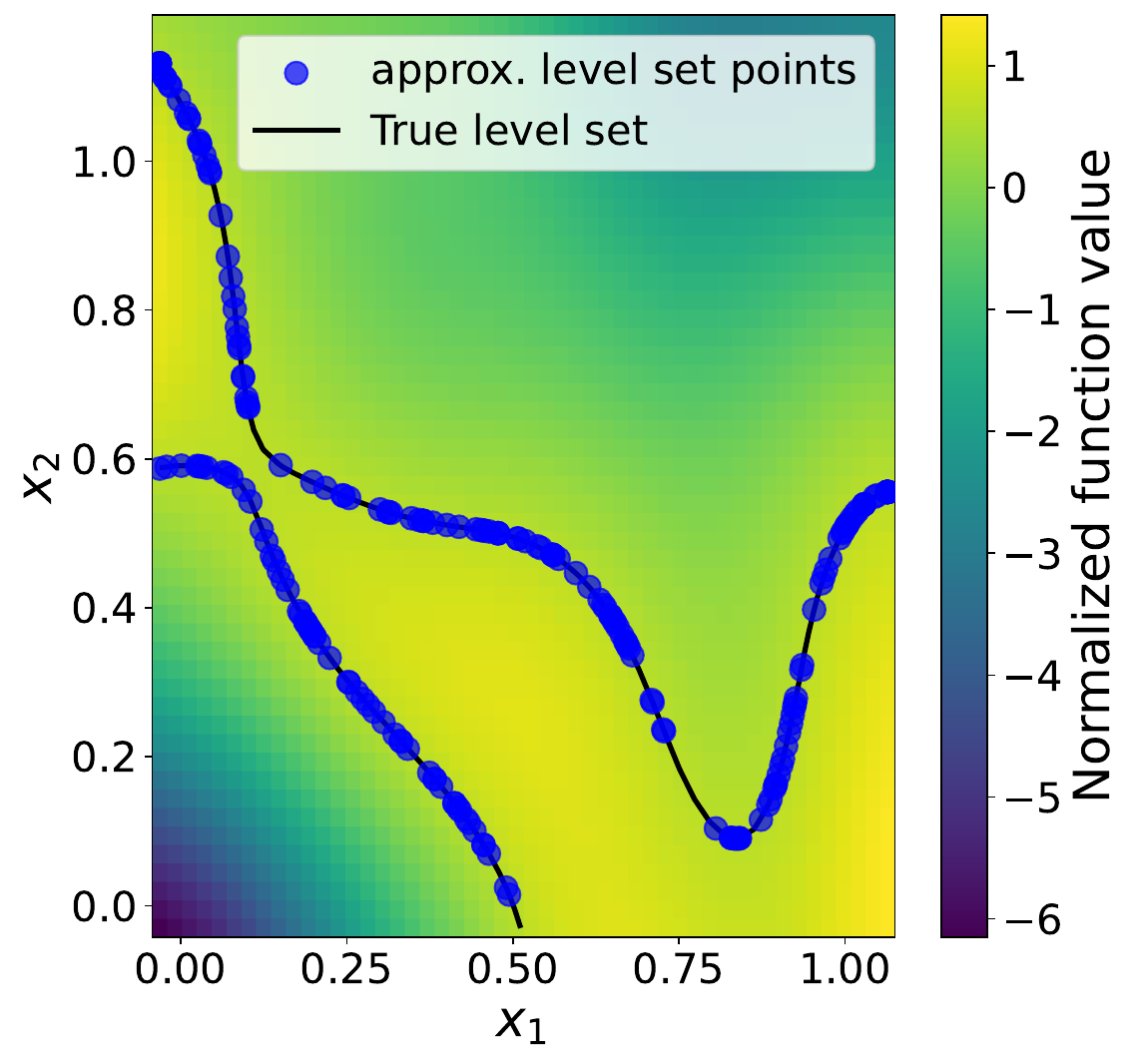}
        \caption{}
        \label{fig:ls}
    \end{subfigure}
    
    \vspace{0.5em}
    
    \begin{subfigure}{0.48\linewidth}
        \centering
        \includegraphics[width=\linewidth]{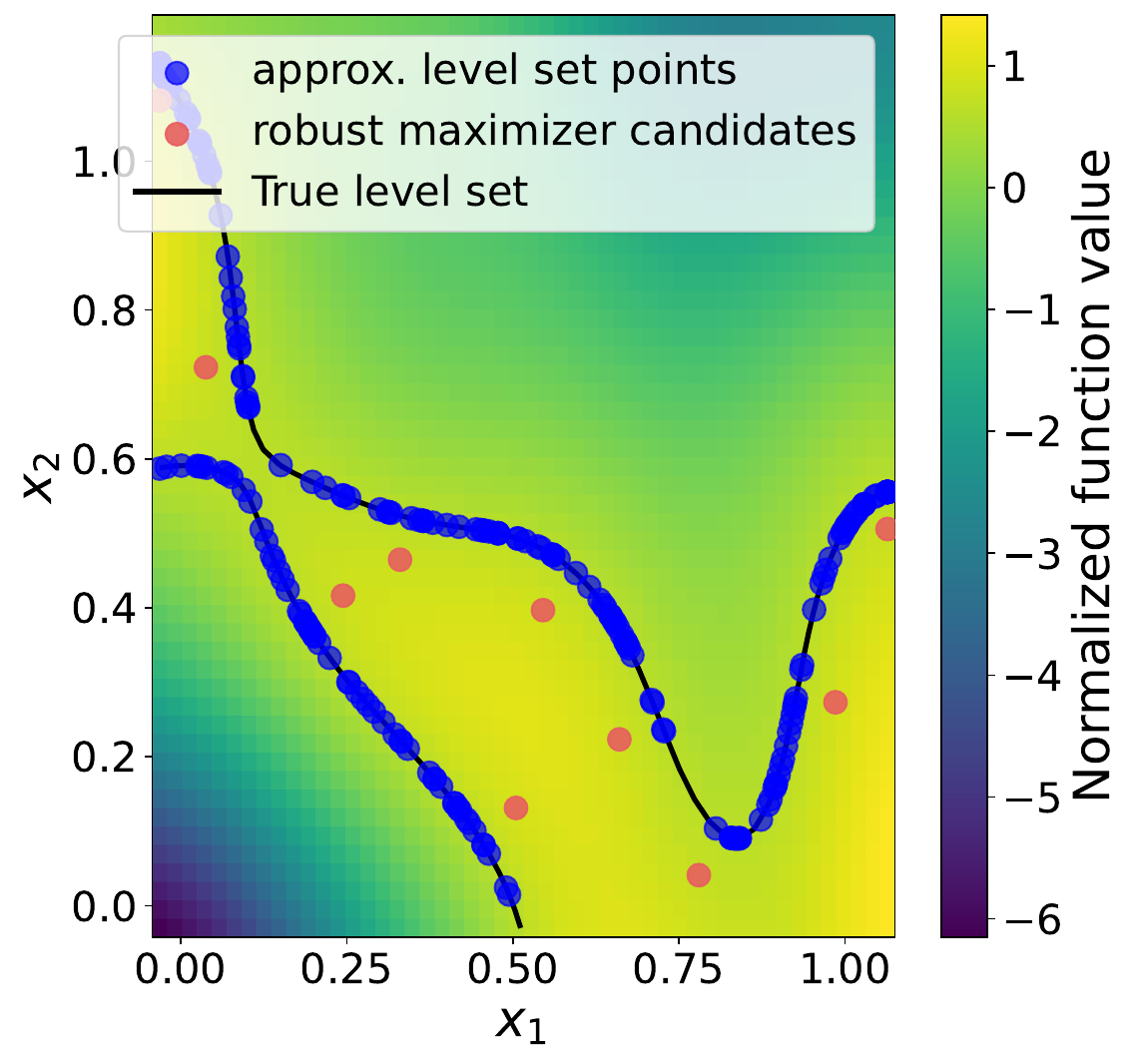}
        \caption{}
        \label{fig:max_init}
    \end{subfigure}
    \hfill
    \begin{subfigure}{0.48\linewidth}
        \centering
        \includegraphics[width=\linewidth]{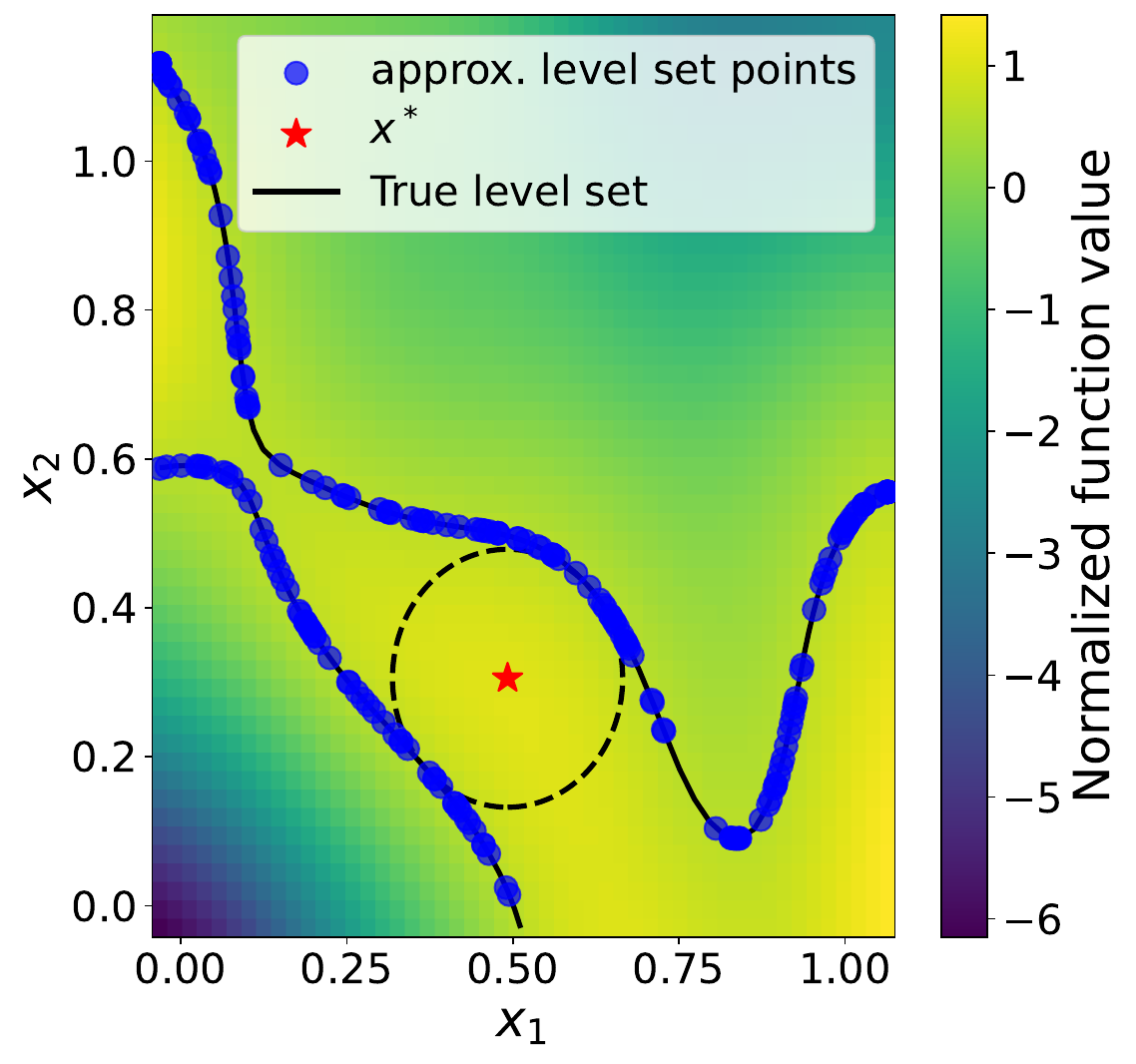}
        \caption{}
        \label{fig:max}
    \end{subfigure}
    
    \caption{Sampling of $\{(x_s^*, r_s^*)\}_{s=1}^S$.
    For each $s$, space-filling initial candidates {\bf(a)} are refined using
    Gauss-Newton to find the level set {\bf(b)}. A robust maximizer {\bf(d)} is found by gradient-based optimization of LogSumExp distance to the level set samples, multi-started from a diverse subset of the level set samples pushed to the superlevel set {\bf(c)}.}
    \label{fig:sampling_comparison_main}
\end{figure}

\paragraph{RFF representation}
By the Bochner's theorem, the spectral density of the GP kernel $k$ exists and we can approximate $k$ as an empirical expectation of the feature inner product $k(x,x') \approx \phi^T \phi$. For the radial basis function (RBF) kernel with the length scales $\ell$ and the output scale $\sigma_f$, $\phi$ is an $m$-dimensional random feature function defined as
\begin{align*}
    \phi(x) = \sqrt{\frac{2\sigma_f}{m}} \cos(Wx + b),
\end{align*}
where the rows of $W\in\mathbb{R}^{m\times d}$ are sampled independently as $w \sim \mathcal{N}(0, I/\ell)$ and the phases are sampled as $b \sim \mathcal{U}(0, 2\pi)$. We approximate the GP prior using a Bayesian linear model
\begin{align*}
    \tilde{f}(x) = \phi(x)^{\top}\theta,
\end{align*}
where $\theta \sim \mathcal{N}(0, I)$ is a Gaussian-distributed weight vector. The posterior 
of 
$\theta|D_t$ is Gaussian,
$
    \theta \sim \mathcal{N}\!\left(A^{-1}\Phi^{\top} y,\; \sigma^2 A^{-1}\right),
$
where $\Phi = [\phi(x_1),...,\phi(x_t)]^{\top}$ is the feature matrix and 
$
    A = \Phi^{\top}\Phi + \sigma^2 I.
$

\paragraph{Robust maxima}
To generate a single approximate posterior sample $f_s$, we draw both the random feature parameters $(w_s, b_s)$ and the corresponding posterior weight vector $\theta_s$ according to the generative process above. This yields a differentiable sample function
\begin{align*}
    \tilde{f}_s(x) = \Phi_s(x)^{\top}\theta_s.
\end{align*}
This enables efficient Algorithm~\ref{alg_continuous_sampling} for finding the robust maximizer
$(x_s^{*}, r_s^{*})$ for each $\tilde{f}_s$. It is visually illustrated in Figure~\ref{fig:sampling_comparison_main}, consisting of four key steps explained next.

First, we construct an estimate of the (inflated) level set
\begin{align*}
    \mathcal{C}_s = \{ x \in \mathcal{X} : |f_s(x) - t| < \tau \},
\end{align*}
for a tolerance $\tau$. The set $\mathcal{C}_s$ is approximated by first sampling $k$ space filling samples $(x_1, \dots, x_k)$ (Fig.~\ref{fig:sampling_comparison_main} (a)), and using the Gauss-Newton algorithm starting from each of these initial samples 
to minimize the squared residual $(f_s(x_i) - t)^2$ until the tolerance is met; see Algorithm~\ref{alg_level_set_approx} in Appendix.
This gives $k$ samples covering the level set (Fig.~\ref{fig:sampling_comparison_main} (b)).

The robust maximizer is then obtained by solving
\begin{align}\label{robust_maximizer_objective}
    x_s^{*} =
    \arg\max_{x \in S_s}
    \left[
    -\frac{1}{\beta}
    \log \sum_{c \in \mathcal{C}_s}
    \exp\bigl(-\beta \, d(x,c)\bigr)
    \right],
\end{align}
where the differentiable objective corresponds to a soft-min approximation of the distance to the level set. We first generate a set of $n \ll k$ superlevel set points $S_s = \{x \in \mathcal{X} : \tilde f_s(x) \ge t\}$ by running a few iterations of gradient ascent for $f_s$ starting from the level set points $\mathcal{C}_s$ (Fig.~\ref{fig:sampling_comparison_main} (c)). The set of $n$ samples are selected to be
maximally diverse via a max--min distance criterion; see 
Algorithm~\ref{alg_diverse_gradient_seeding} in Appendix.

In the final step, $x_s^{*}$ is found as the maximizer of the objective~\ref{robust_maximizer_objective} (i.e. $\text{LogSumExp}(-d(x,C_s))$) by running multi-start gradient ascent starting from the points $S_s$ and picking the best one $x_s^{*}$. This approximates the robust maximizer of $f_s$ (Fig.~\ref{fig:sampling_comparison_main} (d)), and the associated robustness radius is
$
    r_s^{*} = \min_{c \in \mathcal{C}_s} d(x_s^{*}, c).
$

\paragraph{Computational cost}
Even though this algorithm requires an explicit construction of the level set and involves two separate optimization stages, we can use gradient-based methods for all parts and the sampling can be done in parallel for all of the MC samples. For one posterior sample, sampling the approximate level set scales linearly with the number of approximate level set points $k$, basis functions $m$, 
and the dimensionality $d$, with complexity $\mathcal{O}(kmd)$. The computation of the robust maximizer is dominated by evaluating the distances between the $n$ maximizer candidates and the $k$ level set points, which scales as $\mathcal{O}(knd)$. To accelerate this phase when $k$ is large,
we restrict the LogSumExp distance computation in each iteration to the 100 nearest points, since distant points have a negligible contribution. To maintain accuracy, the set of nearest points is re-estimated every 10 iterations. In practice the computational cost is not an issue in typical applications where the function evaluation cost dominates. For example, in a three-dimensional problem the cost for selecting the next candidate was on average 2.4 seconds, only three times the cost of standard BO acquisition (\ref{tab:processing_times}).

\begin{algorithm}[t]
\caption{\textsc{SampleContinuous}}
\label{alg_continuous_sampling}
\LinesNotNumbered
\KwIn{Surrogate GP model $\mathcal{GP}$, number of level set samples $S$, number of basis functions $m$, tolerance $\tau$, threshold $t$}
\KwOut{$\{(x_s^*, r_s^*)\}_{s=1}^S$}
\For{$s = 1,\dots,S$}{
    Construct differentiable sample function $\tilde f_s(x) = \phi_s(x)^\top \theta_s$\\
    Approximate level set samples $\mathcal{C}_s \leftarrow \textsc{LevelSetApprox}(\tilde f_s, t, \tau)$\\
    Initialize robust maximizer candidates 
    $S_s \leftarrow \textsc{InitializeMaximization}(\tilde f_s, \mathcal{C}_s )$\\
    Compute robust maximizer $x_s^* = \arg\max_{x\in\mathcal{S}_s} -\frac{1}{\beta}\text{LogSumExp}(-d(x,C_s))$\\
    Compute robustness radius\\ 
    $r_s^* = \min_{c \in \mathcal{C}_s} d(x_s^*, c)$
}
\end{algorithm}

\subsection{Robustness Regret}
We propose two robustness-based regret metrics for evaluation.
The \emph{expected robustness regret} at iteration $i$ is 
\begin{align}
    \mathcal{R}_i = |\rho(x^*) - \mathbb{E}_{p(\tilde{x}^*|D_t)}[\rho(\tilde{x}_i^*)]|,
\end{align}
where $\tilde{x}^*_i$ denotes the estimated maximizer of Eq. \eqref{eq:x_star_is_rho} at iteration $i$. The expectation is taken over GP sample paths, and the true objective value is evaluated for each sample path. This can be computed for any method, but requires GP approximations similar to what we use in sampling. As a more tractable alternative, we define the \emph{inference robustness regret}, relying only on the GP posterior mean:
 \begin{align}\label{eq_inf_regret}
    \tilde{\mathcal{R}}_i = |\rho(x^*) - \rho(\tilde{x}_i^*)|,
\end{align}
where $\tilde{\rho}_i$ is the posterior mean of $\rho$ at iteration $i$, and $\tilde{x}_i^*$ is the estimate of $x^*$ at iteration $i$, that is $\tilde{x}_i^*=\arg\max_{x \in \mathcal{X}} \tilde{\rho}_i(x)$. This surrogate regret is inexpensive to compute and enables efficient evaluation.

\section{Related Work}

As explained in Introduction, our setup is close to the broad literature on robust BO, especially to worst-case robust optimization, but differs by assuming perturbation-free inputs during optimization. Ignoring this distinction for the moment, the algorithm itself can be contrasted with prior work. AdveRS-2 \citep{saday2025robust} and StableOpt \citep{bogunovic2018adversarially} both rely on the Upper Confidence Bound (UCB) acquisition function, which scales poorly in higher dimensions. StableOpt requires solving an inner optimization problem $\inf_{\delta \in B_r} f(x+\delta) \ge t$ for each candidate point during acquisition optimization. AdveRS-2, in turn, discretizes the design space to identify the point that maximizes the distance to the unsatisficing sub-level set, with exponential cost. We avoid both limitations.
The sampling of the maximizers $(x^*, r^*)$ is independent of the candidate and hence we can first sample the maximizers and only then optimize the acquisition function efficiently using gradients.

Information-theoretic acquisition functions have been used in Expectation Robustness setting \cite{frohlich2020noisy} and in a specific worst-case contextual variable robustness use-case \cite{weichert2024robust}, and have long-standing roots in standard Bayesian optimization \cite{hernandez2014predictive}. Our formulation is closely related to the symmetric mutual-information acquisition functions \citep{hernandez2014predictive, wang2017max}. It also aligns with approaches that jointly consider information about both the maximum value and its location \cite{hvarfner2022joint}, which has been shown to yield more informative acquisitions. Unlike their method, however, ours does not require conditioning the surrogate model on a fantasized maximum, avoiding an additional computational step.

Satisficing BO is intrinsically related to estimation of the level set where $f(x)=t$.
Finding the level set of the true (black-box) function is an extremely hard problem, especially in higher dimensions:
\citet{bachoc2021sample} showed that the sample complexity depends on the packing number of the (inflated) level set, while \citet{gotovos2013active} introduced a GP-UCB-style algorithm with probabilistic convergence guarantees under the assumption that $f$ is drawn from a GP on a discrete domain.
While we need to estimate the level set within the sampling algorithm, it is critically important to note that we do \emph{not} need the level set of the true $f(x)$ but only the level sets of GP samples $f_s(x)$. These samples are smooth, we are not constrained in sample evaluations, and we can use gradients. This makes learning the level set relatively easy for reasonably smooth targets.
Moreover, we do not need perfect coverage of the level set for finding the robust maximizer, as will be empirically demonstrated in Section~\ref{sec:levelset}.

One relevant line of research relates to the nature of the inputs. \cite{frohlich2020noisy} and \cite{ daulton2022robust} consider robustness specifically for controllable inputs, whereas \cite{weichert2024robust, toscano2022bayesian} explicitly assume there are two distinct sets of inputs: design parameters to be optimized and uncontrollable environmental parameters with respect to which the solution should remain robust. Our presentation assumes all inputs will be noisy after deployment, but we can easily extend the approach to accommodate this setting by introducing a binary mask $m \in \{0,1\}^d$ that specifies which coordinates will be perturbed.
This changes the admissible permutation set to
$
B_r(m)
=
\left\{
\delta \in \mathbb{R}^d :
\| m \odot \delta \| < r
\text{ and }
(1-m) \odot \delta = 0
\right\}
$
but requires no other modifications.

\section{Experiments}\label{sec_experiments}
We compare \algabbrev{} with two worst-case robust optimization algorithms and one standard maximization method, explained below, using identical GP surrogate with RBF kernel implemented using \texttt{BoTorch}
\citep{balandat2020botorch} for all methods. The proxy hyperparameters are optimized after each evaluation. Each method is evaluated over 30 independent random initializations, using identical initial designs across all baselines, and the performance is evaluated using the inference robustness regret of Eq.~\ref{eq_inf_regret}, computed with respect to the maximally robust estimate $\tilde{x}^*$ obtained based on the surrogate model. 
See Appendix~\ref{sec_experimental_details} for complete experimental setup. The code for reproducing all experiments is available at \href{https://github.com/KinnunenSamuli/MRSBO/}{https://github.com/KinnunenSamuli/MRSBO/}.

\textbf{Max-value Entropy Search (MES)} \cite{wang2017max} is included as a strong baseline to show how standard optimization algorithms looking for $\arg \max_{x} f(x)$ ignoring robustness of the satisfiability threshold $t$ work in our setting. MES selects points that maximize the mutual information about the unknown maximum value $y^*$ using
\[
\alpha_{MES}(x) 
= H(y \mid x, D_t) 
- \mathbb{E}_{p(y^* \mid D_t)} 
\left[ 
H(y \mid x, D_t, y^*) 
\right].
\]

\textbf{StableOpt} \cite{bogunovic2018adversarially} optimizes worst-case performance under adversarial perturbations with known maximum level $\epsilon$ with UCB acquisition 
\[
\alpha_{SO}(x) = \min_{\delta \in \Delta(x)} \mathrm{UCB}(x+\delta),
\]
where 
$
\Delta(x) = \{x' - x : x' \in \mathcal{X}, \; d(x,x') < \epsilon \}.
$

\textbf{AdveRS-2} \cite{saday2025robust} maximizes our objective but in an adversarial environment. They maximize the optimistic robustness radius under worst-case input perturbations, subject to the constraint $f(x) > t$, with acquisition function
\[
\alpha_{A\text{-}RS\text{-}2}(x) = d(x,x') 
\quad \text{s.t.} \quad \mathrm{UCB}(x+\delta) \ge t.
\]

For both robust baselines, we follow the authors’ recommendations and use $\beta = 2$ to control the UCB exploration level. 
To apply these methods in our setup where the optimization is carried out without adversarial perturbations, we simulate the perturbations by drawing them uniformly within a ball of $\epsilon$-radius.
In absence of prior knowledge the $\epsilon$ parameter could not be determined prior to running the algorithm in realistic settings, but we consider
an idealized setup where we have access to the maximally robust solution $r^*$ and set $\epsilon = r^*$. This can be considered an optimal choice, but we will also illustrate the effect $\epsilon$ has on the methods.

\begin{figure*}
    \centering
    \includegraphics[width=1\linewidth]{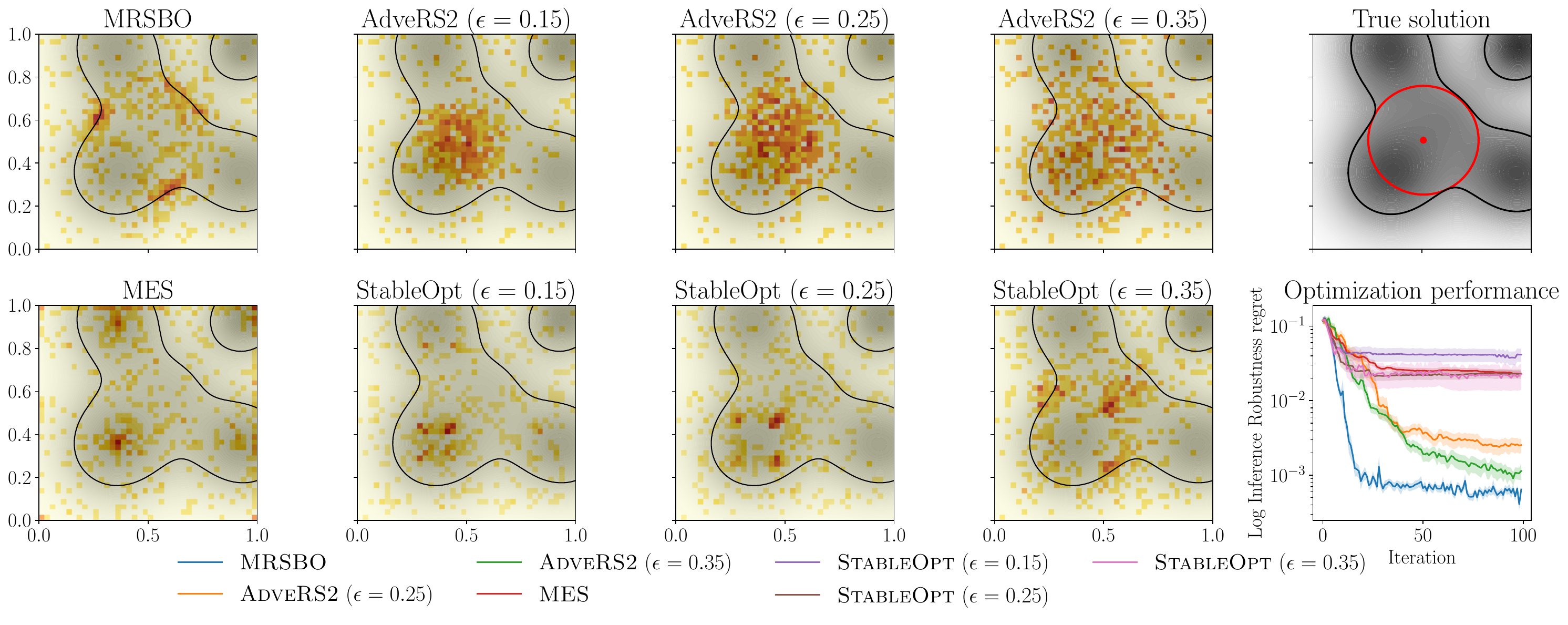}
    \caption{
The first four columns show histograms of the first 20 evaluations aggregated over 30 runs for each acquisition strategy, with the right column showing the solution and regrets. \algabbrev{} directly focuses on evaluations informative of the robustness radius and MES explores the modes to find the global optimum. StableOpt and AdveRS2 correctly explore roughly the area where the robust maximizer is, but AdveRS2 is highly inefficient and StableOpt that ignores the satisficing threshold converges to a wrong solution. For both methods, higher adversarial perturbation $\epsilon$ increases exploration.
}
    \label{fig_experiment_1}
\end{figure*}

\subsection{Benchmarking}
\paragraph{Illustration}
Figure~\ref{fig_experiment_1} illustrates the task and the different ways the algorithms solve it. The target is a two-dimensional Gaussian mixture with multiple modes, showcasing a typical scenario where the global maximizer of the function (at $[0.95, 0.95]$) is vastly different from the true robust maximizer at $x^*=[0.51, 0.51]$. MES obviously finds the global maximizer and hence has substantially worse final regret due to the small robustness radius for that solution, and StableOpt that ignores the satisficing constraint converges to the widest mode -- it is indeed a robust solution, but not optimal for our objective. Both \algabbrev{} and AdveRS-2 converge (approximately) to the robust satiscifing maximizer, but \algabbrev{} achieves it an order of magnitude faster.

To provide insight on how the methods work, we show histograms of the first 20 evaluations, accumulated over the 30 repeated runs. \algabbrev{} spends most of the evaluations exploring the parts of the level set that are informative of the robustness radius, and interestingly often does not evaluate the function at all around the eventual solution. This is in contrast with AdveRS2 that spends most acquisitions learning the function within the superlevel set. MES naturally explores the modes to find the global optimum. StableOpt explores the perimeter of the largest mode, demonstrating somewhat similar evaluation preference as ours.

\paragraph{Low-dimensional targets}
Optimization performance is evaluated on four synthetic benchmarks (Hartmann, Branin, and two Gaussian mixtures) and one real-world task (Robot Pushing~\cite{kaelbling2017learning, wang2017max}). For each function we consider three satisticing thresholds $t$ and the optimization is initialized with 5 random evaluations.

Table~\ref{tab_exp2_regret} shows the regrets for two choices of the threshold $t$ for each data, with evaluation budget $T=20d$; see Appendix~\ref{app_additional_results} for full regret curves and a third (intermediate) threshold.
MRSBO is in general the best solution, reaching zero regret in six scenarios, but AdveRS2 is slightly better in Branin, as well as Robot Pushing for the higher threshold.
The quality of MES depends purely on how good the global maximizer happens to be; it achieves good regret when it coincides with the robust maximizer.

\begin{table}[t]
\centering
\small
\setlength{\tabcolsep}{2pt}
\caption{
Optimization performance (mean and standard error of $\tilde{\mathcal{R}}$) measured with the budget $T=20d$ evaluations, with bold indicating the best method for each task.
See Fig.~\ref{fig_comparison_all_afs} in Appendix for regret curves and exact threshold values.
}
\label{tab_exp2_regret}
\begin{tabular}{lcccc}
\toprule
Method & \textsc{MRSBO} & \textsc{MES} & \textsc{AdveRS2} & \textsc{StableOpt} \\
\midrule
\multicolumn{5}{c}{\textbf{low threshold}} \\
\midrule
Branin & $0.14 \pm 0.04$ & $0.44 \pm 0.08$ & \boldmath{$0.09 \pm 0.02$} & $0.19 \pm 0.04$ \\
GM 2D & \boldmath{$0.00 \pm 0.00$} & $0.03 \pm 0.00$ & $0.01 \pm 0.00$ & $0.04 \pm 0.01$ \\
GM 3D & \boldmath{$0.00 \pm 0.00$} & $0.05 \pm 0.01$ & $0.02 \pm 0.00$ & $0.04 \pm 0.01$ \\
RP 3D & \boldmath{$0.13 \pm 0.03$} & $0.32 \pm 0.06$ & $0.14 \pm 0.03$ & $0.27 \pm 0.07$ \\
Hart. 3D & \boldmath{$0.00 \pm 0.00$} & $0.04 \pm 0.00$ & $0.01 \pm 0.00$ & $0.05 \pm 0.01$ \\
\midrule
\multicolumn{5}{c}{\textbf{high threshold}} \\
\midrule
Branin & $0.06 \pm 0.01$ & $0.09 \pm 0.02$ & \boldmath{$0.05 \pm 0.03$} & $0.13 \pm 0.03$ \\
GM 2D & \boldmath{$0.00 \pm 0.00$} & $0.01 \pm 0.00$ & $0.00 \pm 0.00$ & $0.03 \pm 0.01$ \\
GM 3D  & \boldmath{$0.00 \pm 0.00$} & $0.03 \pm 0.00$ & $0.00 \pm 0.00$ & $0.04 \pm 0.01$ \\
RP 3D & $0.13 \pm 0.03$ & $0.26 \pm 0.04$ & \boldmath{$0.11 \pm 0.03$} & $0.22 \pm 0.04$ \\
Hart. 3D & \boldmath{$0.00 \pm 0.00$} & $0.01 \pm 0.00$ & $0.01 \pm 0.00$ & $0.02 \pm 0.00$ \\
\bottomrule
\end{tabular}
\end{table}

\paragraph{Higher-dimensional targets}
Previous robust BO methods are restricted to targets of very low $d$ due to their inherent max-min optimization structure, often constructing a grid for finding the maximum of $\alpha(x)$ and scaling exponentially with $d$. Our acquisition function is differentiable and we can find the maximum with gradient-based methods as in \cite{wilson2018maximizing}, and hence we avoid this scaling. Constructing the level set for sampling the robust maxima for individual samples intuitively gets harder in higher dimensions, but there is no clear computational bottleneck.

\begin{figure}
    \centering
    \includegraphics[width=1\linewidth]{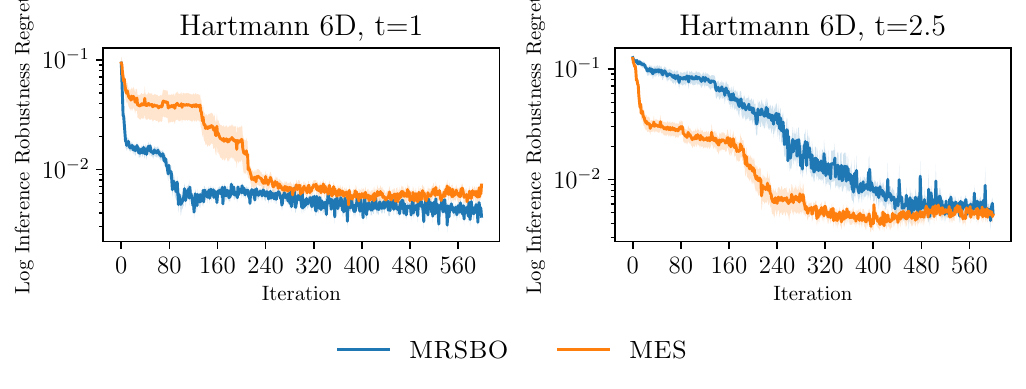}
    \caption
    {Performance on 6-dimensional Hartmann target.}
    \label{fig_experiment3}
\end{figure}
Figure~\ref{fig_experiment3} shows regret curves for a six-dimensional target function (Hartmann 6D) for two thresholds $t$ for \algabbrev{} and MES; StableOpt and AdveRS-2 would no longer scale to this. For sufficiently low threshold \algabbrev{} retains strong performance, converging fast to a good solution. When $t$ approaches the maximum, here $y^*=3.3$, the superlevel set shrinks and all satisficing solutions are here around the global maximizer. MES is then more efficient.
See Appendix~\ref{app_additional_results} for results on other problems of 4-6 dimensions.

\subsection{Level set quality}
\label{sec:levelset}

The quality of the robust maximizer and robustness radius samples $\{x^*, r^*\}_{s=1}^S$ depends on well the level set samples $C_s$ approximate the true level set.  Since Algorithm~\ref{alg_level_set_approx} rejects samples that fail to converge, the accuracy is determined by how well $\mathcal{C}_s$ covers the full level set. For low-dimensional targets we can evaluate the coverage using a dense grid over the space: We identify the cells that intersect the true level set and then measure which fraction of these cells is covered, in the sense that at least one sample in $\mathcal{C}_s$ is in its immediate proximity, up to the grid resolution.

\begin{figure}

      \centering

      \begin{subfigure}{0.48\linewidth}
          \centering
          \includegraphics[width=\linewidth]{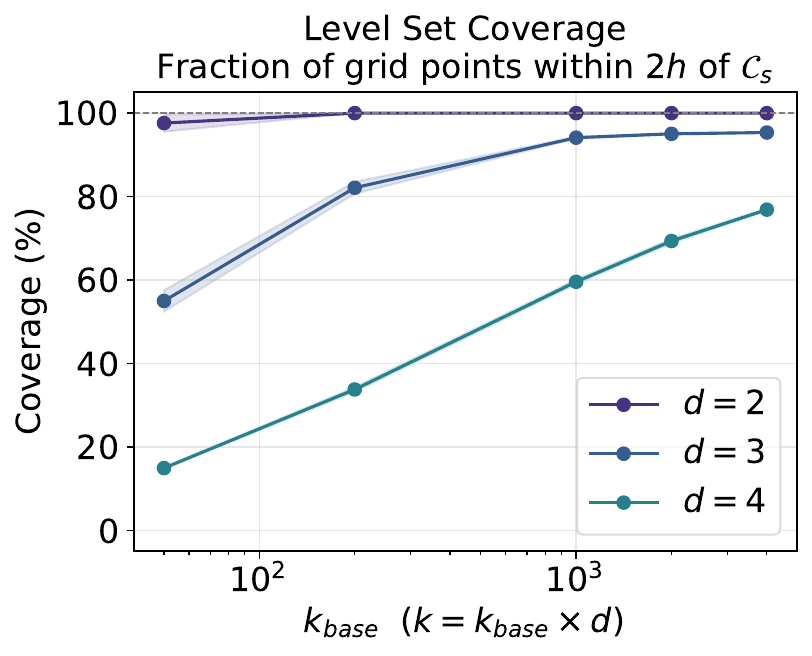}
      \end{subfigure}
      \hfill
      \begin{subfigure}{0.48\linewidth}
          \centering
          \includegraphics[width=\linewidth]{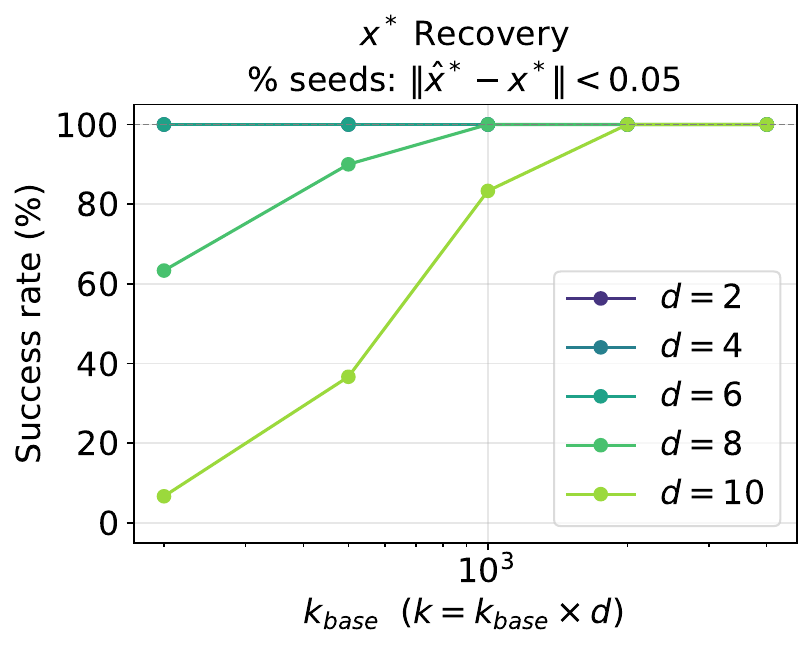}
      \end{subfigure}

      \vspace{0.5em}

    \caption{
\textbf{Left: Level set coverage}. Fraction of the true level set of a bi-modal Gaussian mixture covered by the approximation set $\mathcal{C}_s$. Since the coverage is estimated using a dense grid, the coverage is shown only for $d \le 4$.
\textbf{Right: Robust maximizer identification}. Fraction of robust maximizer estimates $x_s^*$ that converge to the analytic true solution of the bi-modal Gaussian mixture.
    }
      \label{fig_maximizer_recovery_analysis}
  \end{figure}

Figure~\ref{fig_maximizer_recovery_analysis} illustrates the level set coverage and its effect on the robust maximizer identification for a bi-modal Gaussian mixture of varying dimensionality, showing both the level set coverage and the fraction of runs that converge to the correct robust maximizer as a function of $k$, the number of level set samples. The coverage improves with $k$ and degrades with $d$, as expected. Importantly, partial coverage is often enough to find the correct maximizer. For example, with $d=4$ any $k$ is sufficient for finding the correct maximizer even though the coverage is always below $80\%$, and even for $d=10$ (for which coverage cannot even be estimated) we can reliably find the true robust maximizer with large $k$. 
See Appendix~\ref{analysis_of_level_set_coverage} for analysis of the coverage for other targets.

\subsection{Parameter sensitivity}
\begin{figure}
    \centering
    \includegraphics[width=1.0\linewidth]{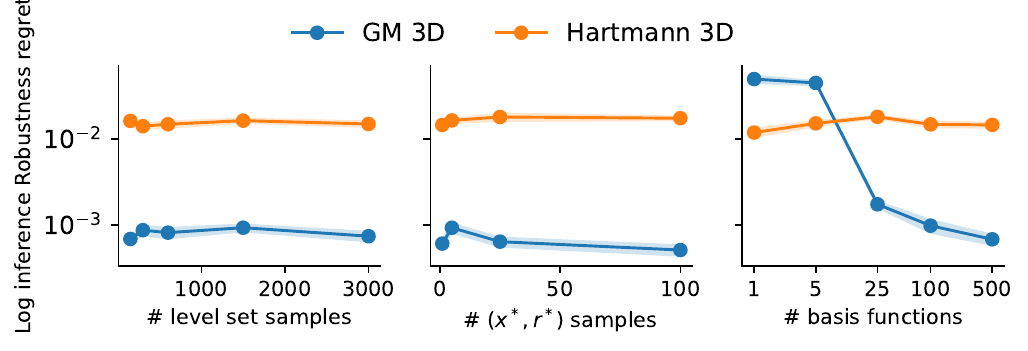}
    \caption{Sensitivity to technical choices: 
    number of level set samples, candidates for robust maximizer, and the basis functions.}
    \label{fig_experiment4}
\end{figure}

Sampling of robust maximizers $(x^*, r^*)$ depends on a few technical hyperparameters. 
Figure~\ref{fig_experiment4} shows how the robust inference regret at budget $T=20d$ depends on these choices for two example targets. We observe that rather small values can be used, speeding up the computation: hundreds of level set samples $k$ and tens of MC samples $S$ are enough in practice, even though we used more in all previous experiments.
However, it is important to use sufficiently many RRF basis functions $m$.

\section{Conclusion}
We introduced a new learning setup for reliable optimization of black box functions when the optimization is carried out in a reliable controlled environment but the eventual solution is deployed in an environment where the inputs may be perturbed. The goal is to find a solution that remains satisfactory, above a given threshold, under maximally large perturbations. We introduced the problem and connected it to the robust Bayesian optimization, explaining how the learning environment is different from previous works. We provided an algorithm that solves the problem using a justified information-gain acquisition, showed how it allocates the evaluations efficiently to explore the level set, and demonstrated the method outperforms standard and robust BO methods in empirical comparisons.

The key technical contribution is the algorithm for sampling robust maximizer candidates for individual GP samples, involving explicit sample-based approximation for the level set. We showed that already rather crude approximations are enough even in dimensionalities higher than typically considered in Robust BO literature, but ultimately any method relying on explicit representation of the level set will fail when the level set is highly complex, i.e. for highly non-smooth target functions in high dimensions. Despite the strong empirical performance, our analysis provides no convergence guarantees or regret bounds; establishing these is an important direction for future work. However, for such problems there are no other good solutions either. Our method is inefficient when the satisficing threshold is very close to the global function maximizer, but for this regime standard BO is a highly practical solution.

\begin{acknowledgements} 
        This work was a part of Finland's Ministry of Education and Culture’s Doctoral Education Pilot under Decision No. VN/3137/2024-OKM-6 (The Finnish Doctoral Program Network in Artificial Intelligence, AI-DOC). This work was supported by ASM International N.V. The authors acknowledge support from CSC – IT Center for Science, Finland, for computational resources. SK, PM and AK additionally acknowledge the research environment provided by ELLIS Institute Finland and support by the Research Council of Finland Flagship programme: Finnish Center
        for Artificial Intelligence FCAI.
\end{acknowledgements}

\bibliography{maximally_robust_satisficing_bayesian_optimization}

\newpage

\onecolumn

\title{Maximally Robust Satisficing Bayesian Optimization\\(Supplementary Material)}
\maketitle

\appendix

\section{Method details}
\subsection{Computational details of the sampling from $p(x^*, r^* \mid D_t)$}

The computational procedure of the sampling algorithm was outlined in~\ref{subsec:sampling}. The method consists of two main stages. First, we construct a differentiable sample path $\tilde{f}_s$ and approximate the corresponding level set. Second, we initialize robust maximizer candidates and compute the robust maximizer.

To approximate the level set (see Algorithm~\ref{alg_level_set_approx}), we construct an estimate of the inflated level set
\begin{align*}
    \mathcal{C}_s = \left\{ x \in \mathcal{X} : |f_s(x) - t| < \tau \right\},
\end{align*}
where $\tau$ is a tolerance parameter. The algorithm begins by initializing candidate level set points using Latin Hypercube Sampling (LHS), which ensures space-filling coverage and promotes a uniform exploration of the domain. 

For each initialized point, we solve a root-finding problem by minimizing the squared residual
\[
    (f_s(x_i) - t)^2.
\]
Each point is optimized independently using the Gauss--Newton method. This choice is motivated by computational efficiency, as it only requires first-order derivatives while providing fast local convergence.

After constructing the approximate level set samples, we select a maximally diverse subset of these points. From this subset, gradient ascent is applied to generate superlevel set points $S_s$, which serve as initialization for the robust maximization step. This initialization procedure is described in~\ref{alg_diverse_gradient_seeding}.

The robust maximizer is obtained by solving
\begin{align}\label{robust_maximizer_objective}
    x_s^{*}
    =
    \arg\max_{x \in S_s}
    \left[
    -\frac{1}{\beta}
    \log \sum_{c \in \mathcal{C}_s}
    \exp\bigl(-\beta \, d(x,c)\bigr)
    \right].
\end{align}

In practice, the dominant computational cost arises from evaluating the distances between the $n$ maximizer candidates and the $k$ level set points, which scales as $\mathcal{O}(k n d)$. When $k$ is large, this step is accelerated by restricting the LogSumExp distance computation to the 100 nearest level set points at each iteration, since distant points contribute negligibly. To preserve accuracy, the nearest-neighbor set is recomputed every 10 iterations.

\begin{algorithm}
\caption{\textsc{LevelSetApprox}}
\KwIn{Sample path $\tilde f_s$, threshold $t$, \# of initializations $n$, tolerance $\epsilon$, damping $\lambda$, max iters $n_{\text{iter}}$}
\KwOut{Approximate level set samples $\mathcal{C}_s$}

Initialize $\mathcal{C}_s = \emptyset$\;

\For{$k=1,\dots,n$}{
    Sample initial point $x^{(0)} \sim \mathrm{LHS}$\;
    
        $x \leftarrow x^{(0)}$\;
    
        \For{$j=0,\dots,n_{\text{iter}}-1$}{
            $r \leftarrow \tilde f_s(x) - t$\;
    
            \If{$|r| \le \epsilon$}{
                Add $x$ to $\mathcal{C}_s$\;
                \textbf{break}
            }
    
            $g \leftarrow \nabla \tilde f_s(x)$\;
            $x \leftarrow x - \dfrac{r}{\|g\|^2 + \lambda}\, g$\;
    }
    }
\label{alg_level_set_approx}
\end{algorithm}

\begin{algorithm}
\caption{\textsc{InitializeMaximization}}
\KwIn{Level set samples $\mathcal{C}_s$, sample path $\tilde f_s$, number of seeds $n$, learning rate $\eta$, ascent steps $T$, 
bounds $(\ell,u)$}
\KwOut{Superlevel set points $S_s$}

Randomly select $c^{(1)} \in \mathcal{C}_s$\;
Initialize $\mathcal{I} \leftarrow \{c^{(1)}\}$\;

\For{$k=2,\dots,n$}{
    For each $c \in \mathcal{C}_s$, compute
    $d(c) = \min_{x \in \mathcal{I}} \|c - x\|$\;
    $c^\star = \arg\max_{c \in \mathcal{C}_s} d(c)$\;
    $\mathcal{I} \leftarrow \mathcal{I} \cup \{c^\star\}$\;
}

$S_s \leftarrow \emptyset$\;

\For{each $x^{(0)} \in \mathcal{I}$}{
    $x \leftarrow x^{(0)}$\;
    \For{$t=1,\dots,T$}{
        $x \leftarrow x + \eta \nabla \tilde f_s(x)$\;
        $x \leftarrow \min\big(\max(x,\ell),u\big)$\;
    }
    \If{$\tilde f_s(x) \ge t$}{
        $S_s \leftarrow S_s \cup \{x\}$\;
    }
}
\Return $S_s$\;
\label{alg_diverse_gradient_seeding}
\end{algorithm}
\subsection{Additional analysis of the level set coverage}\label{analysis_of_level_set_coverage}
\begin{figure*}
      \centering

      \begin{subfigure}{0.48\linewidth}
          \centering
          \includegraphics[width=\linewidth]{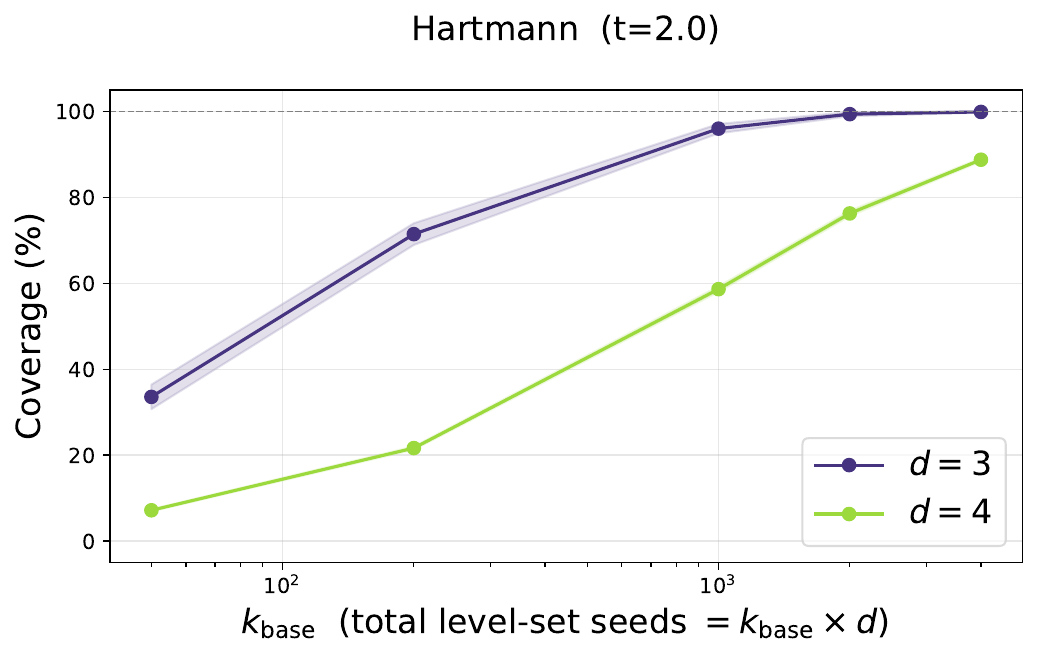}
          \caption{Hartmann}
      \end{subfigure}
      \hfill
      \begin{subfigure}{0.48\linewidth}
          \centering
          \includegraphics[width=\linewidth]{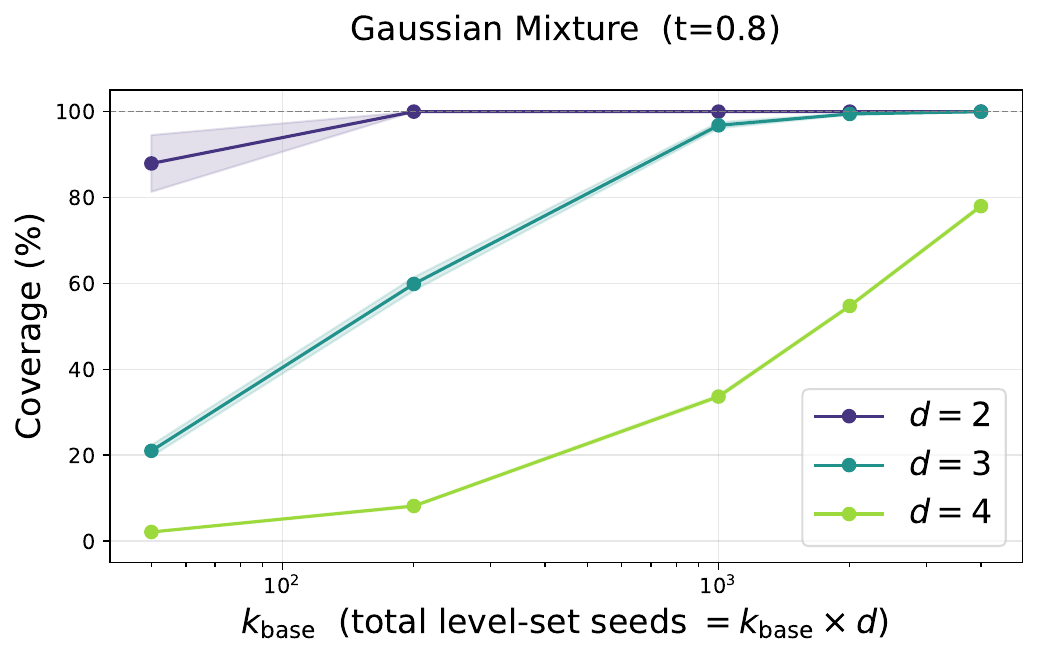}
          \caption{Gaussian Mixture}
      \end{subfigure}

      \vspace{0.5em}

      \begin{subfigure}{0.48\linewidth}
          \centering

  \includegraphics[width=\linewidth]{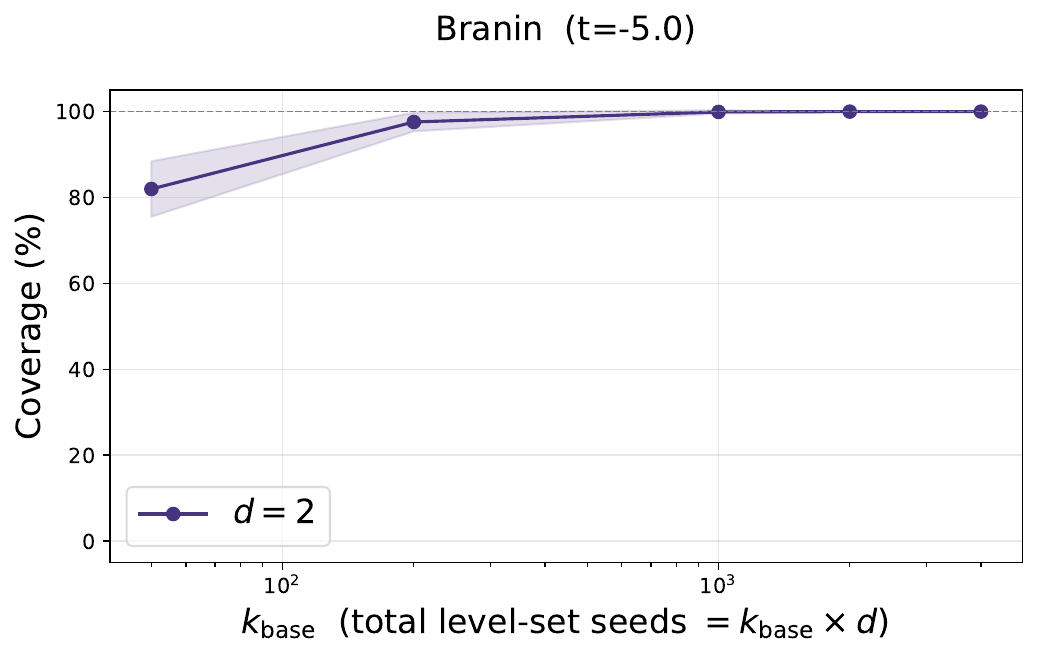}
          \caption{Branin}
      \end{subfigure}
      \hfill
      \begin{subfigure}{0.48\linewidth}
          \centering
      \end{subfigure}
    
      \caption{\textbf{Level set coverage}. Fraction of the true level set, identified as cells of a dense grid that intersect the level set, covered by the approximation set $\mathcal{C}_s$. A cell is considered covered if at least one sample is in its immediate proximity, up to the grid resolution. Covering level sets in higher dimensions is more difficult, but increasing the sample size improves coverage monotonically.}
      \label{fig_level_set_quality_analysis}

  \end{figure*}

The coverage of the approximate level set $C_S$ determines the quality of the robust maximizer and robustness radius samples $\{x^*, r^*\}_{s=1}^S$. We analyze the level set coverage in Fig. \ref{fig_level_set_quality_analysis} for all differentiable synthetic functions used in our experiments up to four dimensions as the analysis requires constructing a dense grid which is computationally infeasible in higher dimensions. The coverage improves with $k$ (the number of level set samples) and decreases with $d$, as expected.

\subsection{Conditional entropy with Lipschitz continuity assumptions}\label{appendix_conditional_with_lipschitz}
Computation of the information gain in Eq.~\eqref{eq_IG_entropy} requires approximating the conditional entropy
$$
H\!\bigl(p(y \mid x, D_t, x^*, r^*)\bigr).
$$
We approximate the conditional entropy by leveraging the property that if the distance to the robust maximizer satisfies $d(x,x^*)\leq r^*$, then by the definition of robustness, function values are constrained to $f(x)>t$. On the contrary, when $d(x,x^*)> r^*$, we do not have such bounds since the function could decrease arbitrarily fast. For sufficiently smooth functions, however, it is possible to construct a lower bound also outside the robustness radius. In the following we discuss one possible approach for doing this, pointing out challenges in obtaining practical gains from such a bound.

\begin{figure}
    \centering
    \includegraphics[width=1.0\linewidth]{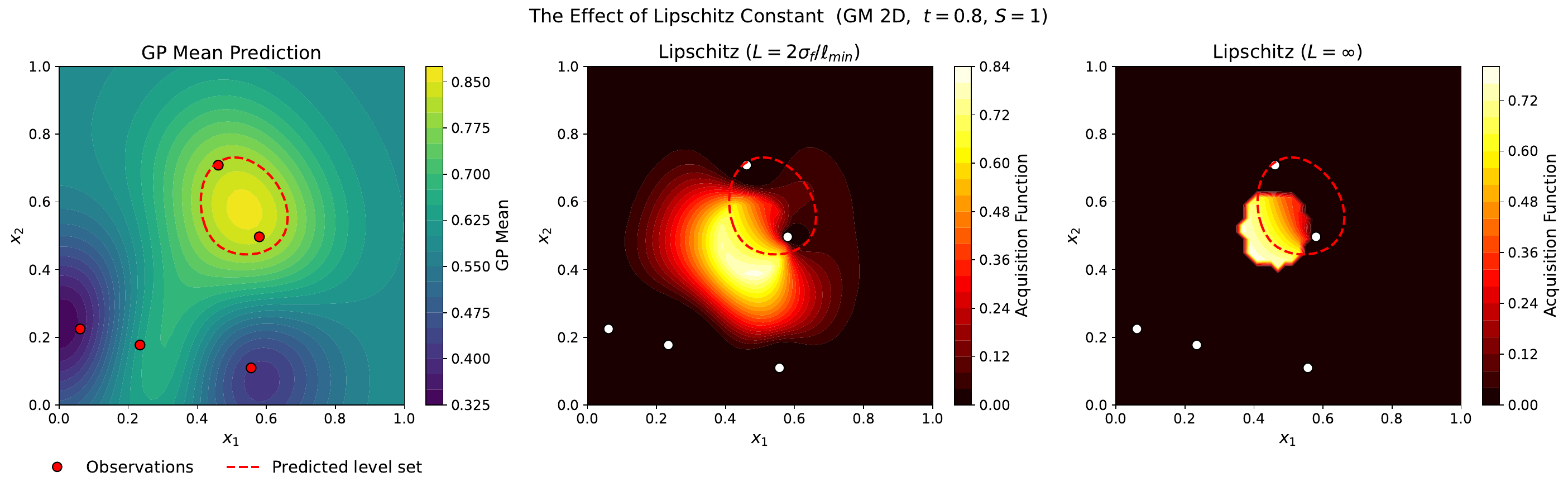}
    \caption{\textbf{Left: GP mean prediction.} The GP posterior mean fitted using 5 observations. \textbf{Middle: $\alpha(x)$ with small $L$.} Acquisition function with $L=2\sigma_f/ \ell_{min}$, where $\sigma_l$ is prior variance and $\ell_{min}$ is the smallest length scale over the two dimensions. The queries expand notably outside of the robustness
radius.  \textbf{Right: $\alpha(x)$ with conservative $L=\infty$.} Without additional smoothness assumption, the acquisitions are more concentrated and more commonly inside the robustness radius.}
    \label{fig_lipschitz_AF}
\end{figure}

If we assume the function is Lipschitz continuous with Lipschitz constant $L\geq0$, we can write a common lower bound that holds both within and outside of the robustness radius as
\begin{align}
&H\bigl(p(y \mid x, D_t, x^*, r^*)\bigr) 
\approx H\!\left(p(y \mid y \ge \kappa(x))\right) \\
&= \log\!\bigl(\sqrt{2\pi e}\,\sigma(x)\, Z(x)\bigr)
+ \frac{\tilde{\kappa}(x)\,\varphi\!\bigl(\tilde{\kappa}(x)\bigr)}{2\,Z(x)}, \label{eq_lipschitz_conditional_entropy}
\end{align}
where $\varphi$ is the probability density function of the standard normal distribution. Here $Z(x) = 1 - \Phi\!\bigl(\tilde{\kappa}(x)\bigr)$ can be computed using the standard normal cumulative distribution function $\Phi$. Now the truncation level $\kappa(x)$ depends on $L$:
\[
\kappa(x) =
\begin{cases}
t, 
& \text{if } d(x,x^*) \le r^*, \\[4pt]
t - L\bigl(d(x,x^*)-r^*\bigr),
& \text{if } d(x,x^*) > r^*,
\end{cases}
\]
used here in a standardized form
$
\tilde{\kappa}(x) = \frac{\kappa(x)-\mu(x)}{\sigma(x)}.
$
Figure~\ref{fig_lipschitz_AF} shows how the acquisition function would change when using  Eq.~\ref{eq_lipschitz_conditional_entropy} for computing the information gain; it encourages the algorithm to sample candidates further outside the current robustness radius. Assuming known $L$, however, is far from realistic and in practice we would need to estimate it.

It is possible, but not easy, to estimate the Lipschitz constant $L$ for a Gaussian process proxy \citep{lederer2019uniform,berkenkamp2023bayesian}. Since we only observe $f$ through a finite set of noisy samples, we do not observe $L$ directly but must infer it.
Moreover, since $f \sim \mathcal{GP}(\mu_t, k_t(x, x'))$ is a random function, the constant $L$ is itself a random quantity: different sample paths of the GP admit different rates of change. Consequently, $L$ can only be estimated up to a probabilistic bound that holds with probability $1 - \delta$. \cite{lederer2019uniform} provides a closed-form estimate $L_f$ of a GP sample $f$
\begin{align}
L_f =
\begin{bmatrix}
\sqrt{2\log\!\left(\dfrac{2d}{\delta}\right)}\,
  \displaystyle\max_{x\in X}\sqrt{k^{\partial_1}(x,x)}
  + 12\sqrt{6d}\,
  \max\!\left\{
    \displaystyle\max_{x\in X}\sqrt{k^{\partial_1}(x,x)},\;
    \sqrt{r\,L_k^{\partial_1}}
  \right\} \\[2.0ex]
\vdots \\[1.0ex]
\sqrt{2\log\!\left(\dfrac{2d}{\delta}\right)}\,
  \displaystyle\max_{x\in X}\sqrt{k^{\partial_d}(x,x)}
  + 12\sqrt{6d}\,
  \max\!\left\{
    \displaystyle\max_{x\in X}\sqrt{k^{\partial_d}(x,x)},\;
    \sqrt{r\,L_k^{\partial_d}}
  \right\}
\end{bmatrix}
\label{eq:lipschitz}
\end{align}
with each element bounding the rate of change along one input dimension.
Here $k^{\partial_i}(x,x') = \frac{\partial^2}{\partial x_i\, \partial x_i'} k(x,x')$ is partial derivative kernel, $L_k^{\partial_i}$ its Lipschitz constant, $r$ the diameter of the domain $X$, $d$ the input dimension, and $\delta \in (0,1)$ the failure probability. All terms are computable analytically for differentiable kernels, but require kernel-specific derivations and the overall estimate depends also on the optimization domain.

\begin{figure}
    \centering
    \includegraphics[width=1\linewidth]{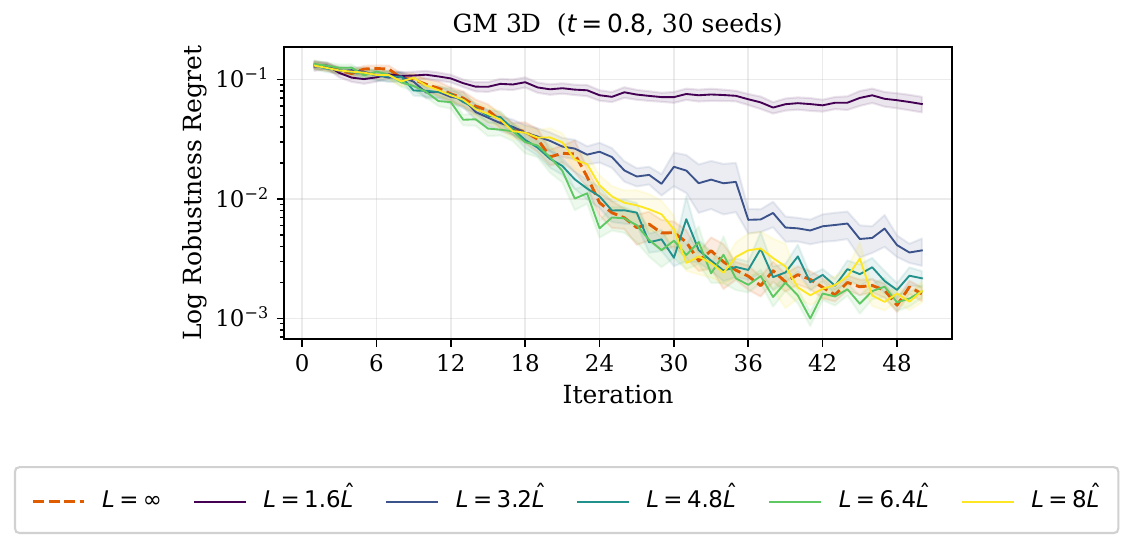}
    \caption{
\textbf{Optimization performance with different Lipschitz constant estimates.} Typical estimators set $L=c \hat L$ for $\hat L = \frac{\sigma}{\ell_{min}}$, where $\ell_{min}$ is the smallest length scale across input dimensions and $c$ depends on several details of the setup. We intentionally consider also overly optimistic estimators with too small $c$, to show the method will then fail. The most pessimistic estimator $L=\infty$ used in all other experiments is highly competitive, but with careful finite estimate (here $c=6.4$) we may gain marginal improvement in early stages of the optimization.}
    \label{fig_lipschitz_regret}
\end{figure}

Instead of constructing a specific estimator, we demonstrate the effect different estimates would have for the actual optimization quality. For the RBF kernel used in our experiments, we have $L \propto \sigma_f/l$ and hence we plot in 
Figure~\ref{fig_lipschitz_regret} the regret as a function of the constant multiplier for this base scale. We see that clearly too optimistic estimators (small $L$) fail completely and for larger multipliers the convergence curves appear highly similar to $L=\infty$ that corresponds to arbitrarily non-smooth function and the computation collapses back to theoriginal formulation in Eq. \ref{eq_conditional_entropy}. With well-chosen $L$ we see very slight improvement over $L=\infty$ during the first 20-25 iterations.

Given the difficulty of estimating $L$ and the marginal gains that can be achieved, we recommend not attempting to bound the function values outside the robustness radius.

\subsection{Conditional entropy with Noisy observations}

In the noiseless-observation case in Section \ref{sec_information_gain}, we used the approximation
\begin{align*}
&H\bigl(p(y \mid x, D_t, x^*, r^*)\bigr) 
\approx H\!\left(p(y \mid y \ge \kappa(x))\right),
\end{align*}
so the conditional distribution becomes a truncated normal with an analytical entropy. In the noisy case this implication no longer holds for the observation \(y\); instead it only applies to the
latent function value $f$.

Following the derivation of the truncation with noisy observations by \citet{takeno2020multi}, we derive \(p(y \mid x, D_t, f>\kappa(x))\) using Bayes' theorem:
\[
p(y \mid x, D_t, f>\kappa(x))
=
\frac{p(f>\kappa(x) \mid y, x, D_t)\, p(y \mid x, D_t)}{p(f>\kappa(x) \mid x, D_t)}.
\]
Here \(p(y \mid x, D_t)\) is Gaussian under the GP predictive model with observation noise, and
\(p(f>\kappa(x) \mid x, D_t)\) is the corresponding tail probability for the latent \(f\).

Assume the observation model \(y=f+\epsilon\) with \(\epsilon\sim\mathcal{N}(0,\sigma_{\epsilon}^2)\),
and the GP predictive for the latent \(f\) given \((x,D_t)\) is \(f\sim\mathcal{N}(\mu,\sigma_f^2)\).
Then the joint marginal is
\[
\begin{bmatrix}
y \\
f
\end{bmatrix}
\sim
\mathcal{N}\!\left(
\begin{bmatrix}
\mu \\
\mu
\end{bmatrix},
\begin{bmatrix}
\sigma_f^2+\sigma_{\epsilon}^2 & \sigma_f^2 \\
\sigma_f^2 & \sigma_f^2
\end{bmatrix}
\right).
\]
Therefore,
\[
f \mid y, x, D_t \sim \mathcal{N}(u,s^2),
\]
with
\[
u = \mu + \frac{\sigma_f^2}{\sigma_f^2+\sigma_{\epsilon}^2}(y-\mu),
\qquad
s^2 = \sigma_f^2 - \frac{\sigma_f^4}{\sigma_f^2+\sigma_{\epsilon}^2}.
\]
Hence,
\[
p(f>t \mid y, x, D_t)=\Phi\!\left(\frac{u-\kappa(x)}{s}\right),
\qquad
p(f>t \mid x, D_t)=\Phi\!\left(\frac{\mu-\kappa(x)}{\sigma_f}\right),
\]
and the Bayes factor is
\[
Z(y)=\frac{p(f>\kappa(x) \mid y, x, D_t)}{p(f>\kappa(x) \mid x, D_t)}.
\]

Finally, the conditional entropy is
\[
H\!\bigl(p(y \mid x, D_t, f>\kappa(x))\bigr)
=
-\int p(y \mid x, D_t, f>\kappa(x))\, \log p(y \mid x, D_t, f>\kappa(x))\, dy.
\]
Using \(p(y \mid x, D_t, f>\kappa(x))= Z(y)\, p(y \mid x, D_t)\), this becomes
\[
H
=
-\int Z(y)\, p(y \mid x, D_t)\,
\log\!\bigl(Z(y)\, p(y \mid x, D_t)\bigr)\, dy,
\]
which can be approximated numerically, e.g. by Monte Carlo with samples
\(y_i \sim p(y \mid x, D_t, f>\kappa(x))\):
\[
H \approx -\frac{1}{N}\sum_{i=1}^N
\log p(y_i \mid x, D_t, f>\kappa(x)).
\]

\subsection{Technical hyperparameters}\label{sub_sec_technical_hypers}
Sampling the robust maximizers $\{(x_s^*, r_s^*)\}_{s=1}^S$ is the most computationally demanding component of our method. Three main technical hyperparameters must be specified. First, the number of Monte Carlo samples $S$ used to approximate the Information Gain in Eq.~\ref{eq_IG_entropy}. Second, to obtain differentiable sample paths, the Gaussian process (GP) surrogate is approximated using Random Fourier Features. The number of basis functions $m$ controls the accuracy of this approximation. Third, for each sampled path, we approximate the corresponding level set. The number of the level set samples $k$ affects the accuracy of the estimated robustness radius: denser samples yield more precise estimates.

\begin{figure*}
    \centering

    \begin{subfigure}{0.48\linewidth}
        \centering
        \includegraphics[width=\linewidth]{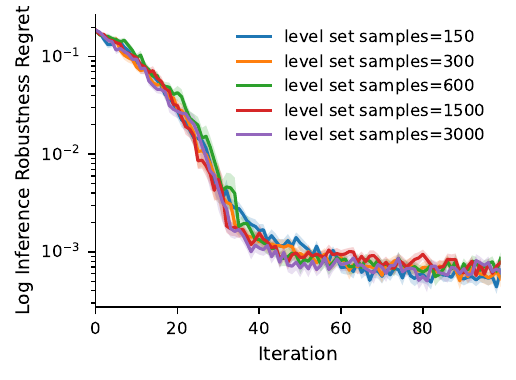}
        \caption{GM 3D $t=0.8$}
    \end{subfigure}
    \hfill
    \begin{subfigure}{0.48\linewidth}
        \centering
        \includegraphics[width=\linewidth]{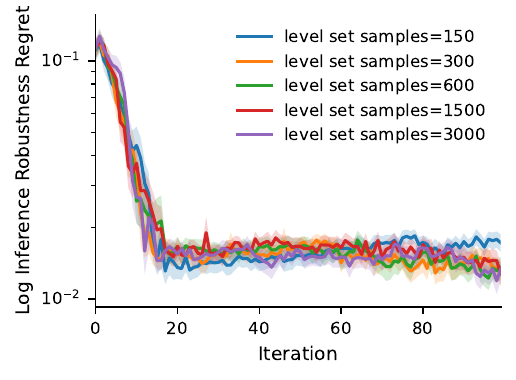}
        \caption{Hartmann 3D $t=2$}
    \end{subfigure}

    \vspace{0.6em}
    \textbf{(a) Sensitivity to the number of  approximate level set samples}

    \vspace{1em}

    \begin{subfigure}{0.48\linewidth}
        \centering
        \includegraphics[width=\linewidth]{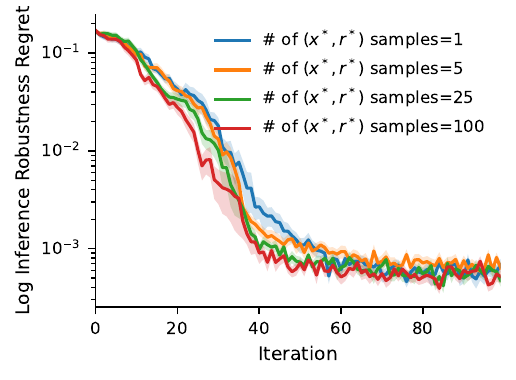}
        \caption{GM 3D $t=0.8$}
    \end{subfigure}
    \hfill
    \begin{subfigure}{0.48\linewidth}
        \centering
        \includegraphics[width=\linewidth]{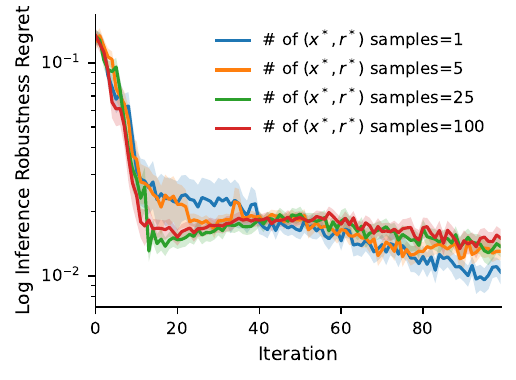}
        \caption{Hartmann 3D $t=2$}
    \end{subfigure}

    \vspace{0.6em}
    \textbf{(b) Sensitivity to the number of robust maximizer samples $(r^*,x^*)$  }

    \vspace{1em}

    \begin{subfigure}{0.48\linewidth}
        \centering
        \includegraphics[width=\linewidth]{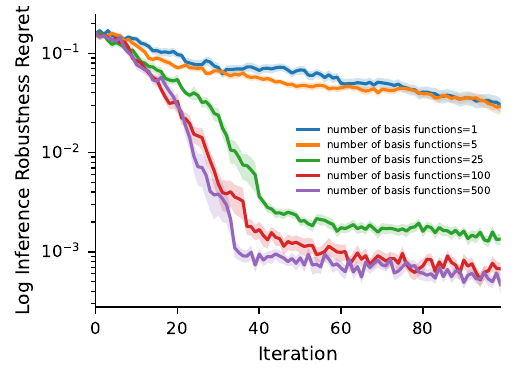}
        \caption{GM 3D $t=0.8$}
    \end{subfigure}
    \hfill
    \begin{subfigure}{0.48\linewidth}
        \centering
        \includegraphics[width=\linewidth]{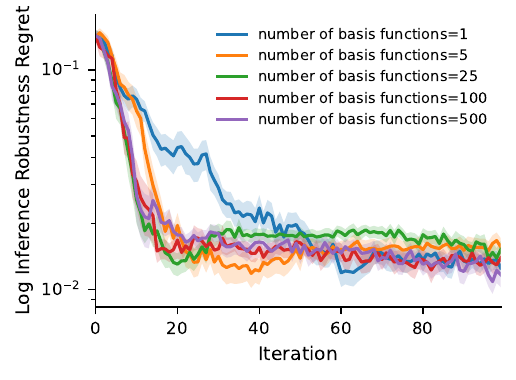}
        \caption{Hartmann 3D $t=2$}
    \end{subfigure}

    \vspace{0.6em}
    \textbf{(c) Sensitivity to the number of basis functions}

    \caption{
    Sensitivity analysis across technical hyperparameters for Gaussian Mixture (GM) 3D (left column) 
    and Hartmann 3D (right column).}
    \label{fig:appendix_sensitivity}
\end{figure*}

A sensitivity analysis of these hyperparameters is presented in Fig.~\ref{fig:appendix_sensitivity}. In this analysis, we vary one hyperparameter at a time while fixing the others to ${S = 100d, m = 100, k = 100}$. Overall, \algabbrev{} is relatively robust to the choice of hyperparameters. The main exception is the number of basis functions. In the Gaussian Mixture (3D) problem, using too few basis functions leads to substantially higher regret.
\subsection{Run time}
In Table \ref{tab:processing_times}, we show average per-iteration runtime for selecting the next candidate in the Gaussian Mixture 3D benchmark. Only the computational cost of acquisition optimization (i.e., selecting the next evaluation point) is included. The time required for GP hyperparameter fitting is excluded, as it is identical across all methods. The runtimes are computed using Intel Core Ultra 5 125U × 14 with 16 GiB of RAM memory. The technical hyperparameters are the same as in the experiments specified in Table \ref{tab:tech_hypers_low_dim}. 

\begin{table}[t]
\centering
\caption{Average per-iteration runtime for selecting the next candidate in the Gaussian Mixture 3D benchmark.}
\label{tab:processing_times}
\begin{tabular}{lcc}
\toprule
Method & Mean [s] & Std [s] \\
\midrule
\textsc{AdveRS2}   & 2.4 & 0.4 \\
\textsc{MRSBO}    & 2.4 & 0.4 \\
\textsc{MES}      & 0.8 & 0.1 \\
\textsc{StableOpt} & 7.8 & 1.5 \\
\bottomrule
\end{tabular}
\end{table}

\section{Experiment details}\label{sec_experimental_details}
\subsection{Implementation}
\label{app_implementation}

\paragraph{Surrogate Model}
We use the same GP surrogate model for all methods:
\[
f(x) \sim \mathcal{GP}\bigl(\mu(x), k(x, x')\bigr),
\]
where the prior mean is constant, $\mu(x) = m$, and the radial basis function (RBF) kernel is defined as
\[
k(x, x') = \sigma_f^2 
\exp\left(
- \frac{1}{2\ell^2}
\sum_{d=1}^{D}
(x_d - x'_d)^2
\right),
\]
where $\sigma_f^2 > 0$ is the output scale, determining the overall magnitude of the covariance, and $\ell > 0$ is the length-scale parameter, controlling the smoothness of the function with respect to the inputs. The term $\sum_{d=1}^{D} (x_d - x'_d)^2$ denotes the squared Euclidean distance between $x$ and $x'$.

The GP hyperparameters include the kernel parameters $m$, $\ell$, $\sigma_f$, and the observation noise standard deviation $\sigma$. After each evaluation, these hyperparameters are fitted by maximizing the log marginal likelihood:
\[
\log p(\mathbf{y} \mid \mathbf{X})
=
\log \int p(\mathbf{y} \mid f(\mathbf{X})) \,
p\bigl(f(\mathbf{X}) \mid \mathbf{X}\bigr)\, df.
\]

For the length-scale and output-scale parameter, we impose a Gamma prior,
\[
\ell \sim \text{Gamma}(\alpha, \lambda),
\]
with shape parameter $\alpha$ and rate parameter $\lambda$ as defined in Table \ref{tab_gamma_priors}.
\begin{table}[h]
\centering
\caption{Gamma priors for the length-scale $\ell$ and output-scale $\sigma_f^2$ for different benchmark functions. The Gamma distribution is parameterized by concentration (shape) $\alpha$ and rate $\lambda$.}
\begin{tabular}{lcccc}
\hline
& \multicolumn{2}{c}{Length-scale $\ell$} & \multicolumn{2}{c}{Output-scale $\sigma_f^2$} \\
\textbf{Function} & $\alpha$ & $\lambda$ & $\alpha$ & $\lambda$ \\
\hline
Branin              & 1.0  & 4.0  & 3.0  & 2.0  \\
Hartmann 3D         & 6.0  & 17.0 & 11.0 & 10.0 \\
Hartmann 4D         & 6.0  & 17.0 & 11.0 & 10.0 \\
Hartmann 6D         & 6.0  & 17.0 & 11.0 & 10.0 \\
GM 3D               & 6.0  & 17.0 & 11.0 & 10.0 \\
GM 4D               & 6.0  & 17.0 & 11.0 & 10.0 \\
GM 6D               & 6.0  & 17.0 & 11.0 & 10.0 \\
Robot Push 3d     & 3.0  & 19.0 & 3.0  & 2.0  \\
\hline
\end{tabular}
\label{tab_gamma_priors}
\end{table}
The GP model is implemented using \textit{SingleTaskGP} from BoTorch~\cite{balandat2020botorch} and GPyTorch~\cite{gardner2018gpytorch}.

\paragraph{Optimization of the Acquisition Function}
For MES and \algabbrev{}, we optimize the acquisition function using multi-start gradient descent~\cite{wilson2018maximizing}using the implementation of BoTorch.

For AdveRS-2 and StableOpt, due to the discretization-based implementation of AdveRS-2 and the nested max--min optimization in StableOpt, we evaluate the acquisition function on an equidistant grid of size $50^d$ in 2D and $30^d$ in 3D, selecting the input that maximizes the acquisition value.

These algorithms were originally developed for adversarial optimization settings, where the maximizer of the acquisition function,
\[
x_t = \arg\max_x \alpha(x),
\]
is adversarially perturbed before evaluation:
\[
y = f(x_t + \delta_t) + \epsilon.
\]

In StableOpt, the perturbation is defined as the worst-case perturbation computed using the GP lower confidence bound $\text{LCB}=\mu(x) - \beta \sigma$:
\[
\delta_t = \arg\min_{\delta \in \Delta_{\epsilon}(x_t)} \mathrm{LCB}(x_t + \delta).
\]

AdveRS-2 considers several perturbation schemes. In our experiments, following the experimental setup of \cite{saday2025robust}, we use random attack perturbations defined as
\[
\delta_t \sim \mathrm{Uniform}\bigl(\Delta_{\epsilon}(x_t)\bigr).
\]

 The perturbation $\delta_{\epsilon} \in \Delta(x)$ is controlled by a stability parameter $\epsilon$. In our experiments, we set $\epsilon$ equal to the admissible perturbation of the maximally robust solution, i.e., $\epsilon = r^*$. However, in real-world settings, determining an appropriate value for $\epsilon$ is challenging, since it requires prior knowledge of the input perturbation levels, which is typically unavailable.

\paragraph{Technical hyperparameters.}
As shown in Section~\ref{sub_sec_technical_hypers}, the proposed method is relatively insensitive to the choice of technical hyperparameters. Therefore, we selected the values primarily according to our computational budget as shown in Tables \ref{tab:tech_hypers_low_dim} and \ref{tab:tech_hypers_high_dim}.

\begin{table}[h]
\centering
\caption{Technical hyperparameters for problems with dimension $d \leq 3$.}
\begin{tabular}{lc}
\hline
Hyperparameter & Value \\
\hline
Number of Approximate level set base samples & 500 \\
Number of  basis functions& 100 \\
Number of $(x^*,r^*)$ samples & 100 \\
\hline
\end{tabular}
\label{tab:tech_hypers_low_dim}
\end{table}

\begin{table}[h]
\centering
\caption{Technical hyperparameters for problems with dimension $d \in \{4,6\}$.}
\begin{tabular}{lc}
\hline
Hyperparameter & Value \\
\hline
Number of Approximate level set base samples & 1000 \\
Number of  basis functions& 100 \\
Number of $(x^*,r^*)$ samples & 100 \\
\hline
\end{tabular}
\label{tab:tech_hypers_high_dim}
\end{table}

\subsection{Robot Pushing benchmark}
The robot pushing optimization problem \cite{kaelbling2017learning} takes a pushing action as an input and outputs the object's distance to the target location. The objective is to minimize the distance to the target location. However, the objective is converted to a maximization problem that takes values $f\in [0, 5]$. The input parameters consist of:
\[r_x \in [-5,5]\]
\[r_y \in [-5,5]\]
\[r_\theta \in [0,30],\]
where $(r_x,r_y)$ denotes the robot location and $r_\theta$ denotes pushing angle.

We use three different Satisficing thresholds $t\in(3, 3.5, 4)$.  

We adapt the open source implementation by \cite{wang2017max} available at \url{https://github.com/zi-w/Max-value-Entropy-Search/tree/master/test_functions}.

\subsection{Hartmann Test Function}

The $d$-dimensional Hartmann function (with $d \in \{3,4,6\}$) is defined on the hypercube $[0,1]^d$. We negate the function to be used in maximization defined as
\begin{equation}
H(\mathbf{x}) 
=  \sum_{i=1}^{4} \alpha_i 
\exp\!\left(
- \sum_{j=1}^{d} A_{ij} \left(x_j - 10^{-4} P_{ij}\right)^2
\right),
\qquad 
\mathbf{x} = (x_1,\dots,x_d)^\top \in [0,1]^d,
\end{equation}
where $\alpha_i \in \mathbb{R}$, $A_{ij} \in \mathbb{R}$, and $P_{ij} \in \mathbb{R}$ are fixed coefficients.

For the six-dimensional case ($d=6$),
\[
\boldsymbol{\alpha} = (1.0,\,1.2,\,3.0,\,3.2),
\]
\[
A =
\begin{pmatrix}
10 & 3 & 17 & 3.5 & 1.7 & 8 \\
0.05 & 10 & 17 & 0.1 & 8 & 14 \\
3 & 3.5 & 1.7 & 10 & 17 & 8 \\
17 & 8 & 0.05 & 10 & 0.1 & 14
\end{pmatrix},
\]
\[
P =
\begin{pmatrix}
1312 & 1696 & 5569 & 124 & 8283 & 5886 \\
2329 & 4135 & 8307 & 3736 & 1004 & 9991 \\
2348 & 1451 & 3522 & 2883 & 3047 & 6650 \\
4047 & 8828 & 8732 & 5743 & 1091 & 381
\end{pmatrix}.
\]

The function is multi-modal. In the six-dimensional setting, it possesses six local minima and a global maximum at
\[
\mathbf{z}^\star 
= (0.20169,\; 0.150011,\; 0.476874,\; 0.275332,\; 0.311652,\; 0.6573),
\]
with optimal value
\[
H(\mathbf{z}^\star) = 3.32237.
\]
We use the open source implementation of Botorch \cite{balandat2020botorch}. Details about the adaptation to the lower dimensional problems available at \url{https://botorch.readthedocs.io/en/latest/_modules/botorch/test_functions/synthetic.html#Hartmann}

We use three different Satisficing thresholds $t\in(1, 2, 3) \text{ if } d\leq4 \text{ else } t\in(1, 2, 2.5)$.  

\subsection{Branin Test Function.}

The Branin  function is a two-dimensional test function
defined on the domain
\[
(x_1, x_2) \in [-5,10] \times [0,15].
\]
It is given by
\begin{equation}
B(x_1, x_2)
=
\left(
x_2 - b x_1^2 + c x_1 - r
\right)^2
+
10(1 - t)\cos(x_1)
+
10,
\end{equation}
where the constants are
\begin{equation}
b = \frac{5.1}{4\pi^2},
\qquad
c = \frac{5}{\pi},
\qquad
r = 6,
\qquad
t = \frac{1}{8\pi}.
\end{equation}

The function is multimodal and possesses three global minimizers,
\begin{equation}
\mathbf{z}_1 = (-\pi,\, 12.275), 
\qquad
\mathbf{z}_2 = (\pi,\, 2.275), 
\qquad
\mathbf{z}_3 = (9.42478,\, 2.475),
\end{equation}
at which the global minimum value is
\begin{equation}
B(\mathbf{z}_i) = 0.397887,
\qquad i=1,2,3.
\end{equation}

We use three different Satisficing thresholds $t\in(-20, -10, -5)$.  

We use the open source implementation of Botorch \cite{balandat2020botorch} available at  \url{https://botorch.readthedocs.io/en/latest/_modules/botorch/test_functions/synthetic.html#Branin} .

\subsection{Gaussian Mixture Test Function.}

The Gaussian mixture test function is defined on the unit hypercube
\[
\mathbf{x} \in [0,1]^d,
\]
with $d \in \mathbb{N}$. It consists of a mixture of Gaussian components, a single mode for each corner and additional mode in the center. Modes are defined as:
\begin{equation}
f(\mathbf{x})
=
\sum_{i=1}^{M}
a_i
\exp\!\left(
-\frac{1}{2}
(\mathbf{x} - \boldsymbol{\mu}_i)^\top
\boldsymbol{\Sigma}_i^{-1}
(\mathbf{x} - \boldsymbol{\mu}_i)
\right),
\end{equation}
where $M$ is the number of modes, $a_i > 0$ are amplitudes,
$\boldsymbol{\mu}_i \in \mathbb{R}^d$ are the component means,
and $\boldsymbol{\Sigma}_i \in \mathbb{R}^{d \times d}$ are diagonal covariance matrices.

\textbf{The mean vectors of the Gaussians} are given by the $2^d$ shifted corner points and one shifted center point. 
Let $\mathbf{v}^{(k)} \in \{0.2,\,0.8\}^d$ for $k=1,\dots,2^d$ enumerate all corner vectors, and let $\mathbf{1}\in\mathbb{R}^d$ denote a vector of ones. 
Define the corner means
\begin{equation}
\boldsymbol{\mu}_k \;=\; \mathbf{v}^{(k)} + 0.15\,\mathbf{1}, 
\qquad k=1,\dots,2^d,
\end{equation}
and define the center mean
\begin{equation}
\boldsymbol{\mu}_{2^d+1} \;=\; 0.5\,\mathbf{1} + 0.15\,\mathbf{1}.
\end{equation}
Thus, the total number of modes is $M = 2^d + 1$.

\medskip
\noindent
\textbf{Base variances.}
All modes are initialized with $\sigma_i^2 = 0.02$ and then modified for selected modes:
the first corner mode is set to $\sigma_1^2 = 1.2 \cdot 0.02$,
the corner mode with the highest amplitude uses $\sigma_m^2 = 0.5 \cdot 0.02 = 0.01$,
and the center mode uses $\sigma_c^2 = 0.01$.

\medskip
\noindent
\textbf{Dimension-dependent scaling.}
The covariance matrices are isotropic and dimension-scaled as
\begin{equation}
\boldsymbol{\Sigma}_i \;=\; \sigma_i^2 \, d^{\alpha} \, \mathbf{I}_d,
\end{equation}
where $\mathbf{I}_d$ is the $d\times d$ identity matrix and
\[
\alpha =
\begin{cases}
1.0, & d=2, \\
0.8, & d=3, \\
0.85, & d=4, \\
0.6, & d=6. \\
\end{cases}
\]

\medskip
\noindent
\textbf{Amplitudes.}
All amplitudes are $a_i=1$ except for the corner mode with the highest amplitude, where $a_m=1.1$, and the center mode, where $a_c=0.4$.

We use three different Satisficing thresholds $t\in(0.7, 0.8, 0.9) \text{ if } d\leq4 \text{ else } t\in(1.0, 1.1, 1.2)$.  

\section{Additional Results} \label{app_additional_results}

\begin{figure*}[t]
    \centering
    \includegraphics[width=\textwidth]{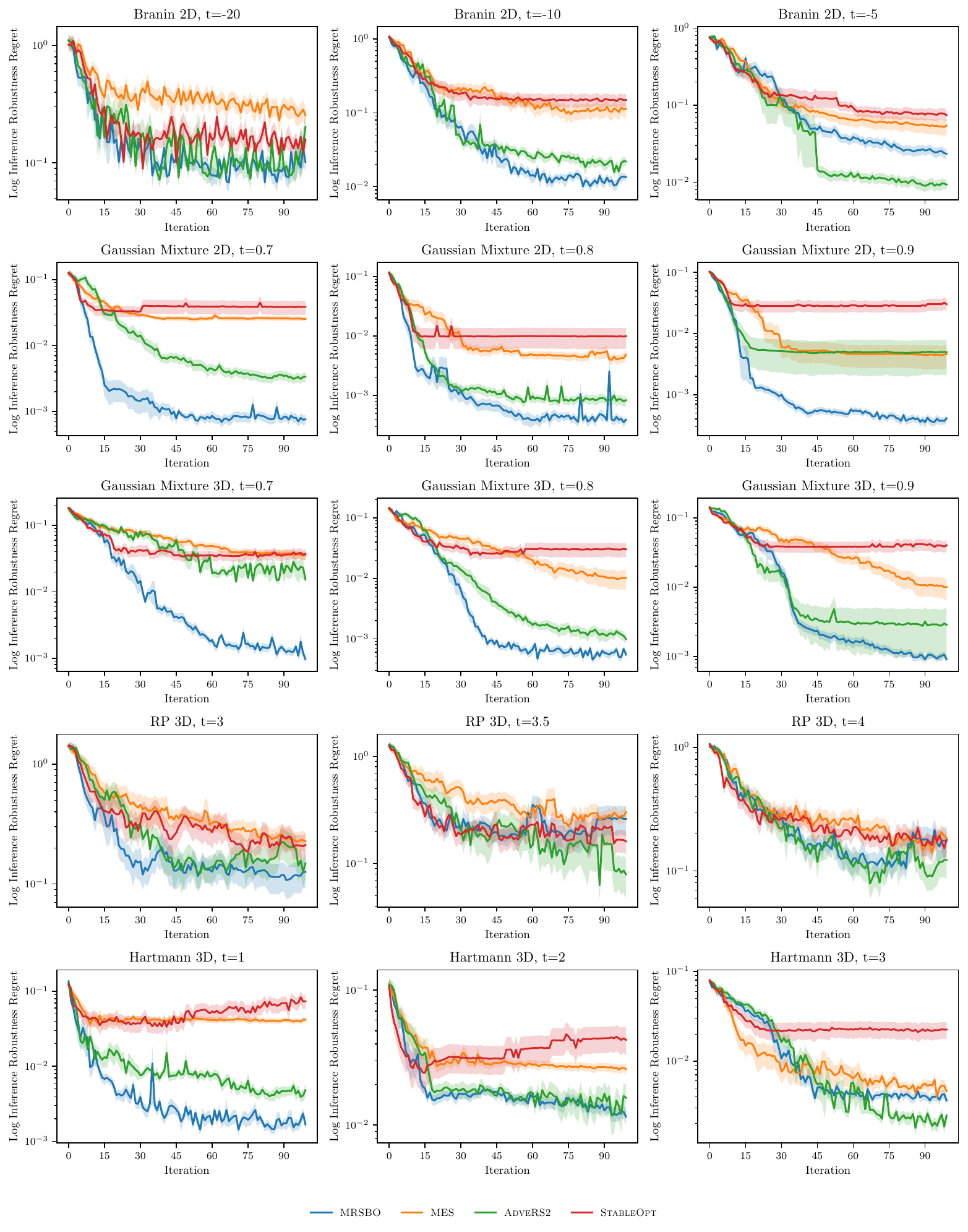}
    \caption{Comparison of AFs $d \leq 3$}
    \label{fig_comparison_all_afs}
\end{figure*}

\begin{figure*}[t]
    \centering
    \includegraphics[width=\textwidth]{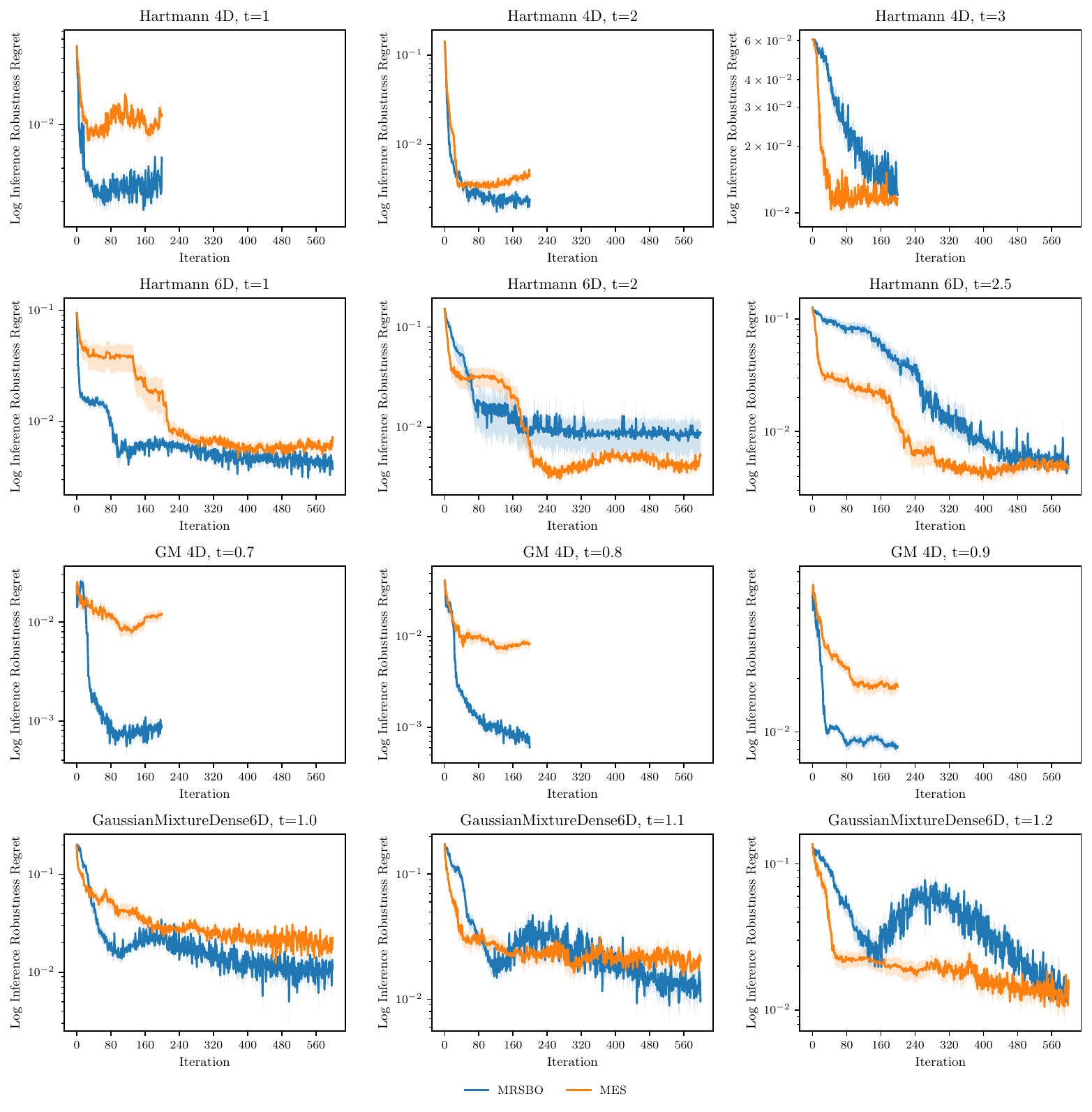}
    \caption{Comparison of AFs 4D and 6D. }
    \label{fig_comparison_high_dim_afs}
\end{figure*}

The complete experimental results are shown in Fig.~\ref{fig_comparison_all_afs} and Fig.~\ref{fig_comparison_high_dim_afs}. For each benchmark and method, we report Robust Inference Regret curves over the full optimization horizon. These results complement the findings presented in Section~\ref{sec_experiments} by providing a more comprehensive view of the empirical performance across all settings.

\begin{figure}[t]
    \centering

    \begin{subfigure}{0.49\linewidth}
        \centering
        \includegraphics[width=\linewidth]{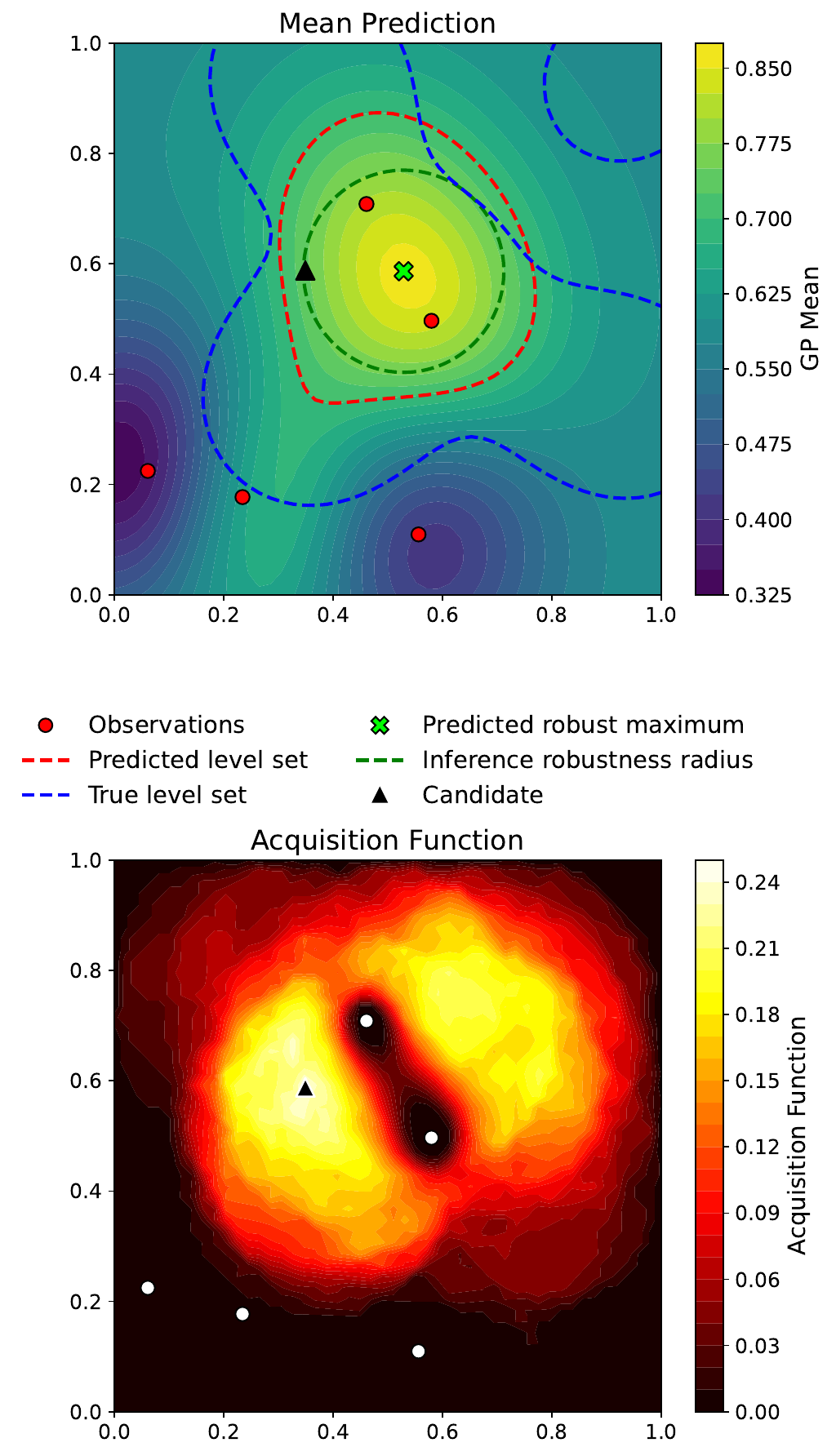}
        \caption{Iteration 1}
        \label{fig:iter1}
    \end{subfigure}
    \hfill
    \begin{subfigure}{0.49\linewidth}
        \centering
        \includegraphics[width=\linewidth]{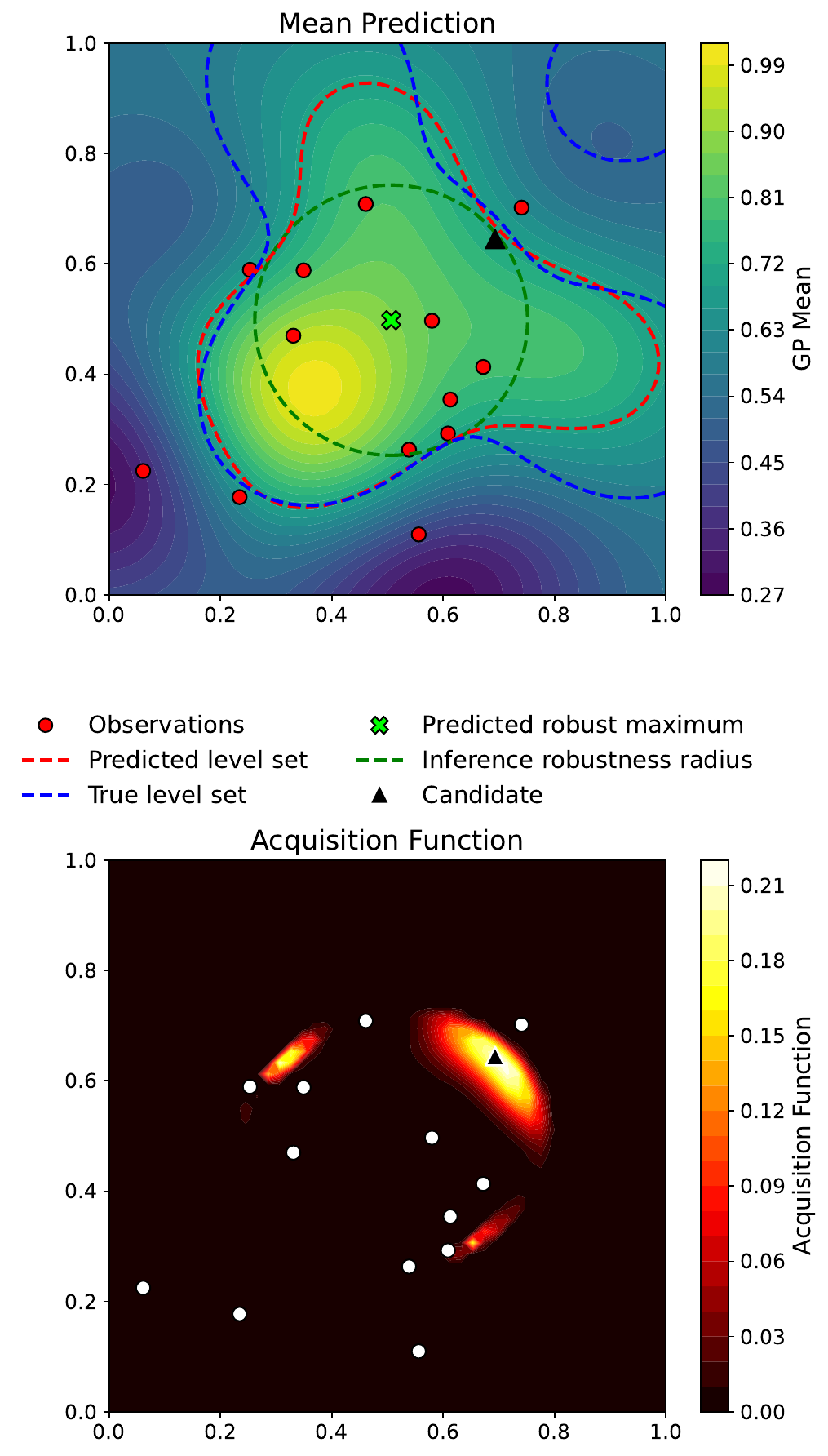}
        \caption{Iteration 9}
        \label{fig:iter9}
    \end{subfigure}

    \caption{
    Progress of \algabbrev{} algorithm in 2-dimensional Gaussian Mixture with $t=0.7$ and random initialization of 5 evaluations.
    The top panel shows the GP mean with predicted level set (red dashed), true level (blue dashed),  inference robust maximizer (green marker), the selected candidate (black triangle), and observations (red points). 
    The bottom panel shows the acquisition function with the selected candidate (black triangle). 
    The model increasingly refines the level set approximation and concentrates exploration near the borders of the estimated maximally robust satisficing region.
    }
    \label{fig:optimization_progress}
\end{figure}

In addition, we include a development view of the optimization process in Figure~\ref{fig:optimization_progress}, illustrating how the proposed acquisition criterion encourages exploration during the early stages and gradually shifts toward exploitation by reducing uncertainty in the relevant regions of the level set.

\end{document}